%% file: main_acl.tex
\pdfoutput=1

\documentclass[11pt]{article}

\usepackage[preprint]{acl-style/acl}

\usepackage{times}
\usepackage{latexsym}

\usepackage{adjustbox}
\usepackage{float}

\usepackage[T1]{fontenc}

\usepackage[utf8]{inputenc}

\usepackage{microtype}

\usepackage{inconsolata}

\usepackage{stfloats}
\usepackage{booktabs}
\usepackage{tabularx}
\usepackage{multirow}
\newcolumntype{L}[1]{>{\raggedright\arraybackslash\hspace{0pt}}p{#1}}

\newcommand*\textmono[1]{\texttt{\expandafter\dottvar\detokenize{#1}\relax}}
\newcommand*\dottvar[1]{\ifx\relax#1\else
  \expandafter\ifx\string-#1\string-\allowbreak\else#1\fi
  \expandafter\dottvar\fi}

\usepackage{footmisc}
\usepackage{enumitem}
\usepackage{caption}
\usepackage{graphicx}
\usepackage{listings} 

\usepackage[capitalize,noabbrev]{cleveref}

\newcommand{\NumBasePersonas}{100 }
\newcommand{\NumBasePersonasSD}{1200 }

\title{Persona-Augmented Benchmarking:\\Evaluating LLMs Across Diverse Writing Styles}

\author{
 \textbf{Kimberly Le Truong\textsuperscript{1}},
 \textbf{Riccardo Fogliato\textsuperscript{2}},
 \textbf{Hoda Heidari\textsuperscript{1}},
 \textbf{Zhiwei Steven Wu\textsuperscript{1,2}}
 \\
 \textsuperscript{1}Carnegie Mellon University,
 \textsuperscript{2}Amazon AWS,
\\
 \small{
   \textbf{Correspondence:} \href{mailto:kltruong@cmu.edu}{kltruong@cmu.edu}
 }
}

\begin{document}
\maketitle
\begin{abstract}
\input{sections/0_abstract}
\end{abstract}


\input{sections/1_intro}
\input{sections/2_related_work}
\input{sections/3_methods}
\input{sections/4_experiments}
\input{sections/5_conclusion}

\input{sections/x_limitations}
\input{sections/x_ethical_considerations}


\bibliography{ref}
\appendix
\input{sections/xx_appendix}




\end{document}

%% file: sections/0_abstract.tex
Current benchmarks for evaluating Large Language Models (LLMs) often do not exhibit enough writing style diversity, with many adhering primarily to standardized conventions. 
Such benchmarks do not fully capture the rich variety of communication patterns exhibited by humans. 
Thus, it is possible that LLMs, which are optimized on these benchmarks, may demonstrate brittle performance when faced with ``non-standard'' input. 
In this work, we test this hypothesis by rewriting evaluation prompts using persona-based LLM prompting, a low-cost method to emulate diverse writing styles.
Our results show that, even with identical semantic content, variations in writing style and prompt formatting significantly impact the estimated performance of the LLM under evaluation.
Notably, we identify distinct writing styles that consistently trigger either low or high performance across a range of models and tasks, irrespective of model family, size, and recency.
Our work offers a scalable approach to augment existing benchmarks, improving the external validity of the assessments they provide for measuring LLM performance across linguistic variations.

%% file: sections/1_intro.tex
\begin{figure}
    \centering
    \includegraphics[width=1\linewidth]{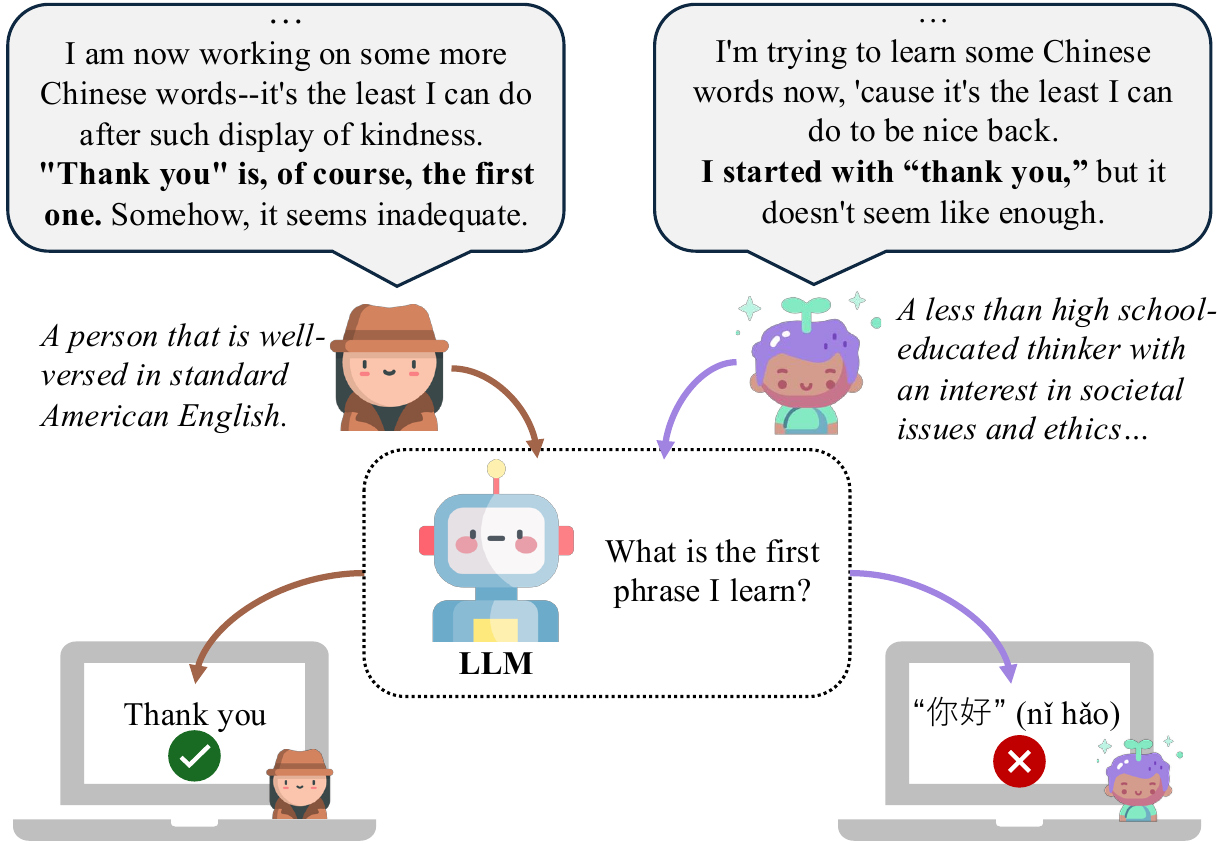}
    \caption{
    A paraphrased example in our experiment. We employ an LLM to rephrase a task in the CoQA benchmark \cite{reddy2019coqa} through two different personas. The evaluated model fails on one rephrasing, despite the answer being entailed in the rephrased text (in bold). See Table \ref{tab:example_coqa} for the full example.}
    \label{fig:motivation}
\end{figure}

\section{Introduction}
Large Language Models (LLMs) have demonstrated remarkable capabilities in a wide range of tasks, yet they exhibit performance disparities when serving different user populations \cite{guo2024benchmarking, arora2025exploring}. This inconsistency stems, in part, from how we evaluate these models. Current benchmarks predominantly feature standardized, formal writing that aligns with dominant linguistic norms found in preprocessed training data \cite{gururangan2022highquality}. By focusing on this narrow linguistic range, these evaluation methods fail to capture the rich diversity of how people communicate \cite{guo2024benchmarking}.
Recent efforts to improve model alignment with human values have unintentionally reduced diversity in LLM outputs \cite{murthy2024fish, reinhart2025llms} with some LLMs lacking the ability to understand diverging writing styles \citep{bhat2025know, shypula2025evaluating}.

In reality, LLMs must consider varying writing styles across dimensions of syntax, lexicon, morphology, and sentiment \citep{biber1991features, biber2009style}.
When faced with inputs deviating from conventional patterns, LLMs exhibit reduced accuracy and inconsistent safeguards based on how information is expressed rather than its content \citep{shi2024llmsafety, grieve2025sociolinguistic}, fundamentally threatening the external validity of benchmark results, where plausible linguistic variations in everyday usage can result in significant variation in performance.

We investigate the impacts of linguistic diversity on LLM benchmarking results by introducing a pipeline for augmenting benchmark datasets to introduce more variation in writing styles. Our approach centers on \emph{personas}--one to three sentence character descriptions combining socio-demographic (e.g., native language, age, education level, gender/sexual identity) and psychosocial (e.g., interests, hobbies, occupation) attributes. These personas guide LLMs to rewrite prompts in ways that reflect how different individuals might express the same information. We refer to the models performing this task as persona-based LLMs. To ensure that the resulting prompts reflect more realistic linguistic variation, our method enforces high-level constraints in the system prompt (e.g., prohibiting the addition of new information, making text understandable to an English-speaking audience to the best of the persona's ability, and providing the option to abstain if the model cannot appropriately rephrase a prompt) while avoiding specific stylistic requirements which may limit linguistic diversity (see full system prompt in \cref{app:prompt_templates}). This controlled approach enables us to evaluate model performance across diverse writing styles without costly human annotation. Using this pipeline, we systematically investigate three key research questions:

\begin{enumerate}
    \item \textbf{Can persona-based LLMs effectively augment existing benchmarks to exhibit diverse writing styles?}
    \vspace{0.1cm}
    \item \textbf{How sensitive are benchmarking results to variations in persona-induced writing style?}
    \item \textbf{Are there writing styles that consistently receive notably high or low performance across a majority of models?}
    \vspace{0.1cm}
\end{enumerate}
\vspace{0.06cm}


\looseness=-1
We demonstrate the effectiveness and scalability of our proposed method on three common benchmarks that evaluate different LLM capabilities: conversational short-answer questions~\citep{reddy2019coqa}, commonsense multiple-choice reasoning~\citep{huang2019cosmos}, and code generation~\citep{lai2023ds-1000}. Our results show that the rephrased prompts exhibit greater linguistic variation compared to the original benchmark, with noticeable stylistic differences between personas. Furthermore, altering a prompt's writing style--while preserving its core content--can significantly impact model performance. Notably, we identify several writing styles that consistently yield high or low performance across all tested models, regardless of model family, size, or release date--even when these same models can correctly identify that the information necessary to answer the question is present in the prompt. One such example is shown in \Cref{fig:motivation}.

These findings highlight two limitations of existing benchmarks: (1) they are often calibrated to a single standardized writing style that poorly represents the diversity of human communication, and (2) they consequently fail to provide externally valid measurements of model performance in real-world applications. 
We offer a practical intermediate solution by providing a scalable approach to improve the external validity of existing LLM benchmarks without requiring costly collection of diverse human data.

\looseness=-1
The consistent performance disparities we identify suggest that even state-of-the-art open-weight models lack robust handling of linguistic diversity, highlighting the need for evaluation methods that capture real-world language variation and for development practices that prioritize writing style robustness. While our work does not claim to definitively reflect authentic human linguistic patterns, it establishes a necessary methodological foundation before conducting more resource-intensive human subject studies to validate these findings in naturalistic settings. Our augmentation method demonstrates value through its diversity, which reveals potential failure modes and enables deeper analysis of the linguistic features and writing styles that models may prefer or struggle with. We observe that the majority of persona-based writing styles yielded worse performance than standard prompts, suggesting potential biases in how LLMs are trained or fine-tuned toward particular linguistic norms. This finding warrants further investigation into training data representativeness and alignment procedures. Future work should optimize persona selection and conduct human-subject experiments to further validate our approach in naturalistic settings.


%% file: sections/2_related_work.tex
\section{Related Work}



\subsection{Writing Style Variation in Benchmarks}
Ideally, LLM evaluation benchmarks should serve as reliable indicators of model quality that enable meaningful comparisons across different models. 
However, these benchmarks have unintentionally encouraged optimization focused on maximizing benchmark scores rather than improving practical performance \citep{mcintosh2024inadequacies, hardt2025emerging}. 
In real-world settings, users interact with LLMs in a wide variety of ways, often expecting models to reason effectively without detailed context and to interpret informal requests \citep{sahoo2024survey, subramonyam2024bridging}. 
However, prominent benchmarks tend to contain 
highly formalized
prompts, resembling standardized tests or technical documentation, rather than reflecting conversational, unstructured, or ambiguous user interactions that are typical of real-world scenarios \cite{ribeiro2020beyond, sarkar2025conversational}. 
Some works have emphasized the importance of multi-prompt evaluations \citep{mizrahi2024state, polo2024efficient}, in which several prompt templates are used for each task; however these require much manual effort to design quality templates and are limited to a fixed set of tasks for each template.
Our work 
introduces a pipeline that can leverage and diversify 
existing benchmarks at scale and without the need for expensive new data collection processes and little human involvement.






\subsection{Sociodemographic Attributes and Writing}
Sociodemographic attributes (e.g., gender, age, education, occupation) are well-established determinants of writing style variation across individuals and communities \citep{koppel2002automatically, sap2014developing, appel2018linking, poole2024llm}. These connections between identity and linguistic expression provide a foundation for generating diverse, authentic writing styles that reflect real-world communication patterns.
Two key studies have examined these effects on LLM prompting. 
\citet{preotiuc2016discovering} analyzed Twitter content, revealing significant differences in syntax (word length, syllables), lexicon (word rareness, vocabulary choice, concreteness), and sentiment expression across demographic groups. 
Their study showed that 
simple lexicon substitution failed to capture the nuanced distinctions between demographic writing styles, highlighting the need for more ``feature-rich'' paraphrasing approaches.
\citet{arora2025exploring} extended this work by using \texttt{Llama2-13b-chat} to rephrase commonsense questions across gender and age groups. Their findings revealed performance degradation for \texttt{Llama2-13b-chat} and \texttt{Mistral-7b-instruct} across more expressive and less formal prompts, with the largest drops occurring for content reflecting younger ages and ambiguous gender. 
However, a methodology that focuses only on isolated demographic features risks over-emphasizing single characteristics rather than holistic personas \citep{hu2024quantifying}--a pattern we also observed in our preliminary experiments.




\subsection{Emulating Writing Styles with LLMs}
Persona-based prompting directs LLMs to adopt specific character roles, simulating how different individuals might express similar information.
While it is clear that the LLM might fail to faithfully represent certain types of users and behaviors~\citep{kapania2025simulacrum}, some positive examples exist. 
\citet{frohling2024personasAttitudes} showed that when personas have specific race and political preferences, LLMs produce toxicity ratings that align with those of comparable human raters. The models also provided similar justifications and stable median ratings across different runs. Similarly, \citet{castricato2025persona} confirmed that LLMs can achieve high inter-annotator agreement with humans emulating the same persona in controversial preference-based tasks.

Persona-based prompting has also been used to improve prompt diversity and model performance \citep{sarkar2025conversational}. 
\citet{chan2024scaling} introduced PersonaHub, a large-scale dataset where personas were obtained by clustering internet metadata. They showed that applying persona-based LLM prompting to rewrite math questions for finetuning increased text diversity and significantly improved performance on standard math benchmarks.

Rather than using personas, 
\citet{cao2024writingStyle} and \citet{errica2025did} emulate writing styles by providing the LLM a brief style guide (one to three sentences) describing variations in writing, such as formality, slang, or emojis when rephrasing benchmark prompts.
Both studies find that the performances of both open- and closed-source LLMs are sensitive to syntactic changes in prompting \citep{sclar2024quantifying, zhuo2024prosa, mizrahi2024state}.  
While these approaches create more diverse prompts, \citet{arora2025exploring} found that explicitly giving a style guide provides worse demographic alignment than just specifying some group and allowing the model to infer and places a ceiling on how much text diversity can be created. 
Our work addresses these limitations by incorporating both psychosocial and sociodemographic information when defining personas and avoiding stringent writing guidelines.



%% file: sections/3_methods.tex
\section{Our Benchmark Augmentation Pipeline}

\begin{figure*}[ht]
    \centering
    \includegraphics[width=1\linewidth]{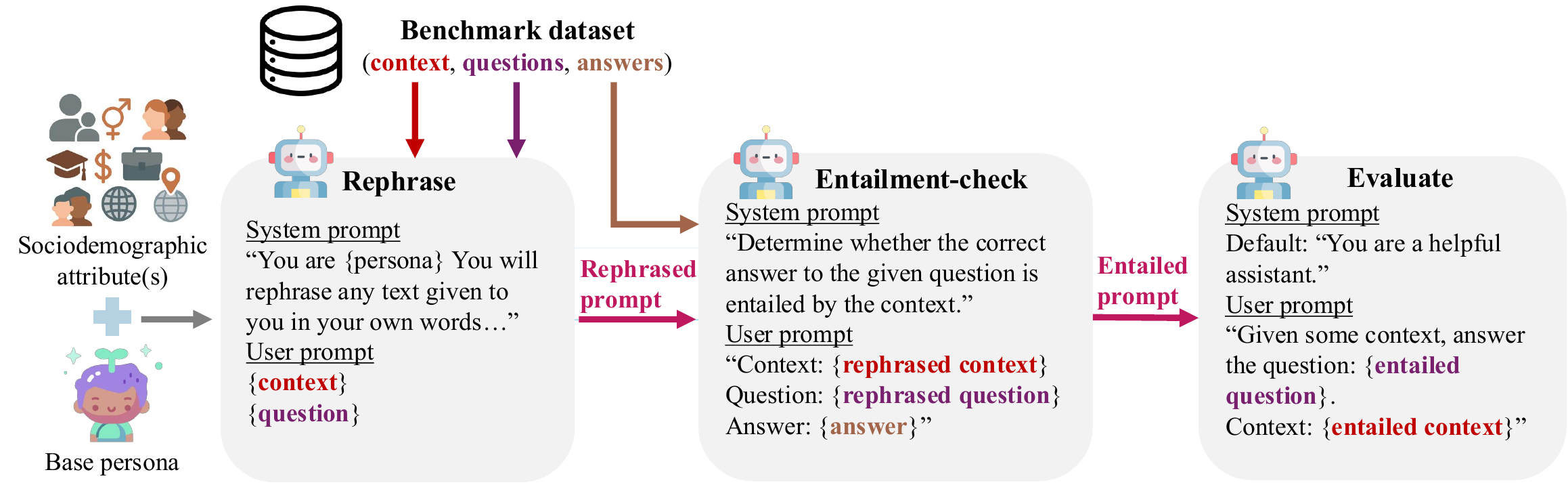}
    \caption{Overview of our methodology to augment benchmark datasets with test instances of the format: context, one or more questions, and one or more correct answers. (1) Initializing personas, (2) Rephrasing the benchmark prompts (contexts and questions), (3) Filtering out question-answer pairs that the LLM refused to rephrase or are not entailed by the rephrased contexts, and (4) Using those contexts to evaluate some LLM.}
    \label{fig:full_methods_diagram}
\end{figure*}

We introduce a pipeline to measure how variations in writing style affect LLM performance. Specifically, we compare a model's performance on prompts rephrased by a persona-based LLM to our baseline: prompts that are rephrased in Standard American English (SAE).
Our proposed method for augmenting existing benchmarks consists of four key steps (see \Cref{fig:full_methods_diagram}): (1) creating a set of persona descriptions containing both sociodemographic attributes (e.g., age and education) and psychosocial attributes (e.g., occupation and hobby), (2) rephrasing benchmark examples using these personas, (3) entailment-checking to ensure the preservation of the original prompt’s content, and (4) evaluating models using the rephrased examples. 
System and user prompts for all processes are in \Cref{app:prompt_templates}.

Our pipeline applies to any benchmark dataset structured as: context, questions, and ground truth answers.
In our experiments, we focus on benchmarks where each example includes context longer than three sentences because it allows us to better analyze how writing style is associated with model performance. In essence, our pipeline simulates scenarios in which a user encounters a problem but is uncertain where to find an answer. In such cases, the user turns to an LLM assistant, providing contextual information about their issue along with a question. Both the context and the question reflect the user’s persona; in subsequent sections, we refer to this pair as a prompt.

\subsection{Choose the Personas}
We design personas characterized by varying psychosocial attributes (e.g., interests, occupation, hobbies) and sociodemographic attributes (e.g., gender/sexual identity, native language, education level, age) rather than explicitly defining linguistic features. This approach aims to elicit diverse writing styles from the LLM while avoiding the reinforcement of stereotypes or the overemphasis of any single attribute.

We select base personas from the PersonaHub dataset \cite{chan2024scaling} to cover the psychosocial elements. To increase diversity, we first randomly select one base persona, then iteratively add personas that maximize the number of distinct n-grams ($n = 4$) \cite{damashek1995gauging} in the set of persona descriptions. 
We manually review these base personas to ensure that their description does not contradict potential sociodemographic attributes that will be added in the next step. 
For example, a persona described as ``\emph{A neurologist who specializes in the study of Parkinson's disease, particularly the mechanisms underlying the development of the disease in different populations and the potential environmental causes},'' would be unlikely for someone with limited education or of a younger age.
We then augment these base personas with four types of sociodemographic attributes: native language (Chinese, English, Spanish), gender/sexual identity (male, female, LGBTQ+)\footnote{More precise terminology would distinguish between gender identity (cisgender male, cisgender female, etc.) and sexual orientation. Our use of ``LGBTQ+'' as a category alongside ``male'' and ``female'' represents a limitation resulting from our experimental design.}, highest education level (less than high school-educated, high school-graduate, college-graduate), and age range (teenager, adult, elderly). We append each individual attribute to the description of the base persona. Thus, in total we have 12 variations (4 attributes $\times$ 3 values each) of every base persona. Adding sociodemographic attributes to the set of personas results in more variation than increasing the amount of base personas (\Cref{app:fig:base_personas_variation_boxplot}).
We also label each persona for whether it contains positive, neutral, or negative connotations about the persona's character. 
See \Cref{app:persona_init} for the exact template and the final personas chosen. 

\subsection{Rephrase Benchmark Examples}
We prompt LLMs with personas to rewrite benchmark test examples in each persona’s writing style while preserving the original meaning.  
We specifically instruct the LLM to maintain all the information contained in the original prompt, to ensure that the written example is understandable to an English-speaking audience, not to add any additional information, and simply refuse to answer if it is not confident that it can properly rephrase the context.

\subsection{Check Entailment of Rephrased Examples}
We then verify that the modified examples retain all necessary information for accurate question answering. Initially, we manually reviewed 60 of the most extreme examples  (i.e., those with the most substantial performance change from the original benchmark performance) for each benchmark. We then employ an LLM directly in our pipeline to determine whether the rephrased context entails all the ground-truth answers to the associated questions. To minimize errors, we evaluate the entailment capabilities of various LLMs and select those with the lowest false positive rates. We prioritize minimizing false positives over false negatives, as this conservative approach excludes examples lacking essential information, even if it means unnecessarily discarding some valid rephrased prompts. As a result, any performance degradation we observe represents a lower bound on the true impact of writing style variation, since all retained prompts are confirmed to preserve the necessary information. We confirm this intuition by employing a second entailment checking LLM. Using these two models, we obtain two ratings for entailment. If the models disagree on a particular example (i.e., one claims the answer is entailed by the context and the other does not), then we add this to the ``disagreement region.'' We use samples within this region to construct error bars for all our performance estimates to account for any biases exhibited by the entailment-checking LLM (see the full procedure in \Cref{app:entailmentCheckErrors}).

\subsection{Evaluate LLM Performance}
\looseness=-1
We perform two corrections for sampling bias that may result from entailment-checking.
Each persona retains a different subset of entailed prompts due to question complexity, possible LLM rephrasing errors, and difficulties adapting to various writing styles. We account for sampling bias resulting from the former two events by: (1) weighting each persona by the number of prompts it successfully entailed when calculating average performance across personas, and (2) stratifying the original benchmark by question difficulty. Specifically, we use k-means clustering ($k=10$) on the average performance of each original prompt across all evaluation models. Then we apply a standard post-stratified estimator \citep{cochran1977sampling} to re-balance difficulty across the entailed set. See \cref{app:poststratify} for the full correction procedure.

We systematically evaluate LLM performance across three versions of each benchmark: original examples, SAE-rephrased examples, and persona-based rephrased examples. To isolate the effects of introducing diverse writing styles from a model having poor rephrasing abilities, we use both the original and SAE-rephrased examples as baselines. This design allows us to distinguish between performance degradation due to a poor rephrasing model (by comparing original to SAE-rephrased examples) and the effects of persona-induced writing styles (by comparing persona-based rephrased examples with SAE-rephrased examples).

%% file: sections/4_experiments.tex
\section{Experiments}
In this section, we describe our experimental setup and report empirical results from
applying persona-based augmentation to three benchmarks.
%
We focus on the key details of the setup; the full description is relegated to \Cref{app:experimental_design}.

\paragraph{Benchmarks} We evaluate the models on three benchmarks, which cover a range of LLM use-cases including factual understanding, common sense reasoning, and code generation.
\begin{itemize}
    \item Conversational Q\&A (CoQA) \cite{reddy2019coqa}: Short-answer questions based on conversational text. We use all 500 examples (each with 10 to 25 questions) from the validation set to demonstrate how writing style affects fact extraction. We report results based on both recall (i.e., token-level overlap of the answer with correct gold labels as reported by \citet{reddy2019coqa}) and cosine similarity (considering the semantics between the answer and correct gold labels).

    \item Common sense Q\&A (CosmosQA) \cite{huang2019cosmos}: Common sense questions (e.g., inferring about human behaviors, intentions, and social interactions) with four multiple choice answers each. We use the full test set of 5675 examples (each with one to four questions) and estimate accuracy to demonstrate how common sense reasoning is affected by writing style. 

    \item Data science code generation (DS-1000) \cite{lai2023ds-1000}: Python code generation tasks given some data science questions written in natural language. We use the full test set of 1000 examples (each with one to two questions) and estimate accuracy to demonstrate that LLMs may fail at programming and logical reasoning due to the writing style.
\end{itemize}

\begin{figure*}[ht]
    \centering
    \includegraphics[width=1\linewidth]{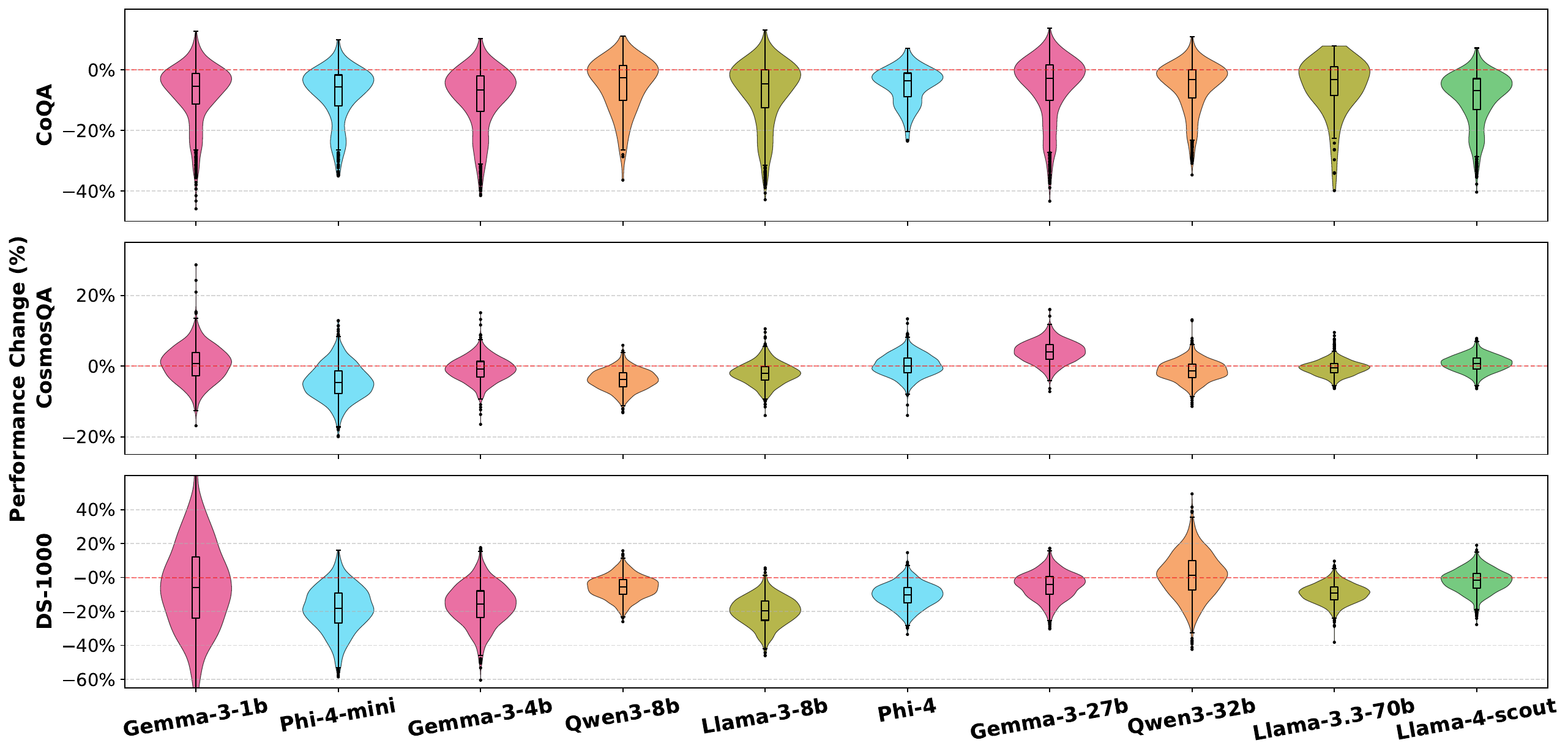}
    \caption{The performance changes (\%) when using persona-rephrased prompts compared to the original benchmark across ten LLMs on three tasks: CoQA (top), CosmosQA (middle), and DS-1000 (bottom). Each violin plot represents the average performance difference received compared to SAE rephrasing by some LLM. The performance of all models are sensitive to writing styles with performance changes varying by 15-80\% \protect\footnotemark \;for a single model across different persona subsets. 
    }
    \label{fig:overall_percent_change}
\end{figure*}


For all experiments, we compare LLM performance on rephrased prompts to two baselines: performance after rephrasing in SAE and performance on the original benchmark without rephrasing. We consider \NumBasePersonas base personas each with 12 sociodemographic attributes, resulting in a total of \NumBasePersonasSD rephrased variations of each original prompt.

\paragraph{Entailment LLMs}
\looseness=-1
We use two open-weight LLMs, \texttt{Gemma-3-27b-it} and \texttt{Qwen-3-32b}, for rephrasing and entailment checking. This procedure produces two sets of rephrased prompts and ensures that any observed variation is not due to model selection. Both models show strong alignment with each other when determining prompt entailment, with approximately 85\% overlap across all three benchmarks. We consolidate results and show figures for \texttt{Gemma-3-27b-it} since it was more conservative in accepting entailment. To assess the sensitivity of the entailment model in our experimental setup, we use the 15\% of cases where the models disagree on entailment to form our disagreement region, which is then used to construct error bars around our performance estimates. We find that across different subsets of personas, the worst performing personas (i.e., personas receiving performance in the 20th percentile) remain the same. Including any fraction of the disagreement region leads to worse results approximately 88\% of the time. Thus any performance degradation reported in our paper is a conservative estimate. Additional details on these sensitivity analyses are provided in \cref{app:entailmentCheckErrors}.

\paragraph{Evaluation LLMs}
On the rephrased prompts, we evaluate the performance of ten LLMs for each set of experiments, covering four model families with varying model sizes and release dates: \texttt{Gemma-3} (\texttt{1b-it}, \texttt{4b-it}, \texttt{27b-it}) \citep{team2025gemma}, \texttt{Llama-3} (\texttt{8b-instruct}, \texttt{70b-instruct}) \citep{grattafiori2024llama}, \texttt{Llama-4-scout-17b-16e-instruct} with 109b parameters \citep{meta2025llama4scout}, \texttt{Phi-4} (\texttt{mini-instruct}, \texttt{-instruct}, with 4B and 14B parameters respectively) \citep{abdin2024phi}, \texttt{Qwen-3} (\texttt{8b}, \texttt{32b}) \citep{qwen3}. In subsequent sections, we refer to these models by their family name and size.

\section{Findings}\label{sec:results}
Using our persona-augmented benchmarks, we investigate the linguistic diversity in the rephrased prompts and whether the linguistic patterns are associated with LLM performance. Specifically, we analyze the best- and worst-performing personas, where performance is measured across all tasks using the average of the performance metric (e.g., cosine similarity for COQA; accuracy for CosmosQA and DS-1000). Best-performing personas are those in the highest quartile of overall performance across all tasks, while worst-performing personas are those in the lowest quartile.

\footnotetext{\texttt{Gemma-3-1b} on DS-1000 is likely an outlier due to very low performance of $\approx3\%$ accuracy even on the original benchmark.}

\paragraph{RQ1: Persona-augmented benchmarks exhibit more linguistically diverse writing styles than both SAE rephrasings and the original benchmark texts.}
Though prior works found that LLMs tend to produce homogeneous text in open-ended generation tasks (e.g., essays, long-form QA, code generation) \cite{alvero2024authorship, reinhart2025llms, shypula2025evaluating}, we find that LLM-rephrased prompts exhibit substantial linguistic variation when rephrasing with strict guidelines, i.e., instructed to maintain semantic content.

To assess textual diversity we measure the average distinct n-grams (across $n=2$ to $n=5$) and within-dataset cosine similarity between prompts. For a fair comparison with the original benchmark, we create a balanced subset of our augmented benchmark by randomly selecting one persona-based rephrasing per original prompt to match the original benchmark size. We repeat this process 25 times. We find that our persona-augmented subset exhibits greater linguistic variation and reduced repetition of common phrases compared to the original benchmark as evidenced by higher distinct n-gram scores ($0.84$ compared to $0.75$ with standard error $<0.01$). Additionally, persona-based rephrasing increases the linguistic variation between independent prompts with lower average within-dataset cosine similarity between test instances ($0$ compared to $0.11$ with standard error $<0.01$).

Furthermore, we verify persona-based rephrasing produces meaningfully different writing styles, while preserving semantic content by conducting a prompt-level analysis. For each original prompt, we measure the pairwise cosine similarity between all of its persona-generated variations. This analysis yields an average cosine similarity of 0.66 between different rephrased versions of the same prompt, indicating that there exists substantial stylistic variation between prompts sharing the same core information.

\paragraph{RQ2: Benchmark results are sensitive to linguistic variation.}
\label{para:A2}
\looseness=-1
Figure \ref{fig:overall_percent_change} demonstrates the impact of persona-emulated writing styles' on LLM performance, with variation ranges of 27-55\%, 9-46\%, and 43-80\% across CoQA, CosmosQA, and DS-1000 respectively. \texttt{Phi-4} uniquely maintains stability across all tasks, while our largest models (\texttt{Llama-3.3-70b} and \texttt{Llama-4-scout}) show pronounced sensitivity exceeding 45\% on CoQA, despite five smaller models possessing less performance variance.

While these aggregate ranges highlight overall model sensitivity, analyzing specific persona subsets provides more granular insights into writing style variations. Comparing best- and worst-performing persona subsets reveals average performance differences of 20, 11, and 28 percentage points for CoQA, CosmosQA, and DS-1000 respectively (\Cref{app:fig:percent_change_trend}). \texttt{Qwen-3-32b} exemplifies these effects with reductions from 0.14 to 0.12 (cosine similarity), 0.78 to 0.73 (accuracy), and 0.18 to 0.12 (accuracy) across benchmarks (\Cref{app:fig:overall_diff}). Some persona subsets trigger performance decrements up to 35\% compared to SAE rephrasing.

Sensitivity to linguistic variations appears to be correlated with task complexity. The most challenging benchmark, DS-1000 (with $59\%$ accuracy from state-of-the-art models\footnote{\label{fn:leaderboard}The official leaderboard can be found at https://ds1000-code-gen.github.io}), exhibits the widest performance spread, while the least challenging one, CosmosQA, shows little variation. This may suggest that commonsense reasoning is less sensitive to linguistic variation compared to conversational question-answering and code generation tasks.

We further examine how increasing writing style diversity affects the benchmark rankings. We hypothesize that writing style differences could potentially lead to instability in benchmark rankings in competitive leaderboards. Despite relatively stable rankings in our experiments (with only minor shifts, see \Cref{fig:performance_compare_w_rank})--likely stemming from our deliberate selection of models with widely varying capabilities--the magnitude of observed performance shifts (5-28 percentage points) could significantly impact rankings in scenarios with more closely matched models. Our simulations using the DS-1000 leaderboard\footref{fn:leaderboard} indicate potential rank changes from -19 to +14 positions depending on the distribution of persona types used in evaluation (\Cref{fig:ds1000_leaderboard_ranks}).\footnote{This simulation assumes other models' rankings would remain unchanged relative to their performance shifts.} In such competitive environments, even a five percentage point performance shift could alter a model's ranking by up to 16 positions. Similar patterns emerged in our CoQA and CosmosQA simulations.


Given that just a fraction of a percentage point often determines benchmark rankings, this instability undermines the validity of current benchmarking practices and suggests that many performance differences may reflect sensitivity to writing style rather than true capability differences. 
\paragraph{RQ3: Certain personas are consistently associated with drops in performance across all evaluated LLMs regardless of the model family, size, recency, or task type.} 
We believe this performance degradation stems from inherent biases in LLM training data and fine-tuning processes that favor standardized, formal writing patterns. All rephrased contexts were prompted to follow correct American English conventions, contain no grammatical errors, and use common vocabulary. These were later confirmed to contain these features through some manual inspection and entailment checking, yet performance varies significantly. (See an example in \cref{app:example_rephased}.) Furthermore, we observe strong Pearson correlation between the model performances received by individual personas for CoQA ($r = 0.84$) and DS-1000 ($r = 0.44$), indicating that when one model performs poorly or strongly on a specific persona, other models tend to show similar performance patterns on that same persona for factual QA and code generation tasks. In contrast, CosmosQA shows minimal correlation ($r = 0.07$), reinforcing our finding (RQ2) that task difficulty and sensitivity to writing style may be correlated.

\looseness=-1
Furthermore, approximately 7-20\% of personas trigger consistently poor performance across at least 6 of 10 models in all tasks. Notably, CoQA and DS-1000 contain 10 and 3 personas, respectively, which under-perform for all 10 models. Across all three tasks, 14 personas consistently rank among the worst performers for at least 6 models--we refer to these as ``global worse-performing personas.'' These performance drops occur independently of model size, family, or recency. As shown in \Cref{fig:overall_percent_change}, our largest model, \texttt{Llama-4-scout}, exhibits poor performance with nearly all identified global worst-performing personas, actually covering more than its predecessors: \texttt{Llama-3-8b} and \texttt{Llama-3.3-70b}. Similarly, despite being used for both rephrasing and entailment \texttt{Gemma-3-27b} also performs poorly for 12 out of 14 personas on all three benchmarks. 


\textit{What linguistic patterns characterize the worst-performing personas?}
Certain sociodemographic attributes consistently affect performance across models and tasks. By grouping personas according to attributes like native language, education level, age, and connotation, we find that education level and native language strongly correlate with performance patterns in CoQA and CosmosQA, while education level, age, and sentiment show the strongest correlations for DS-1000. The presence of these sociodemographic attributes influences performance variation substantially more than base personas. For example, adding different attributes to a single base persona results in accuracy differences up to 0.12, while variation across 100 different base personas is only 0.09 (see \Cref{app:fig:base_personas_variation_boxplot}).

\looseness=-1
Performing a quantitative analysis of the linguistic features also reveals clear stylistic differences between high- and low-performing personas. We computed several linguistic metrics averaged across all models with prompts from both SAE rephrasing and the original benchmark. We find that the best-performing personas tend to use more academic and technical language, with higher Flesch readability (at a early-college grade level) \citep{flesch1948new, kincaid1975derivation}, more nouns indicating more concrete and information-dense text, and more complex sentence structures \citep{biber2009style}. Conversely, the worst-performing personas demonstrate middle-school level readability, simpler sentence structures, and higher hedges (i.e., more words that showed uncertainty). Full results with metric definitions are in \Cref{app:analysisWritingStyles}.

To further characterize these differences, we measured within-dataset cosine similarity for the best- and worst-performing personas. We find that the most common writing style among the best personas can--in nearly all cases--be defined by their base persona. In contrast, the worst-performing personas are overwhelmingly defined by specific sociodemographic attributes injected into them--most commonly ``less than high school-educated'' or age descriptors such as ``teenager'' or ``elderly.'' While high-performing personas typically have positive or neutral character connotations, the worst-performing personas have the same distribution of connotations found in the base persona set, suggesting that sociodemographic attributes overshadow connotation effects for poor performers.

Consolidating our findings across models and tasks, we identified 14 personas (shown in \Cref{tab:all_worst_personas}) that consistently ranked in the bottom quartile for at least 6 of 10 models on all three benchmarks. Sociodemographic attributes drive performance variation much more than the base personas themselves. Of the 14 personas, 9 are described as ``less than high school-educated'' and 4 are described as ``elderly.'' We also found the base personas that received poor performance contained cultural characteristics associated with rural or isolated settings (e.g., ``\emph{English native speaker from a small town who has not traveled much}''), more vulnerable identities (e.g., an ``\emph{elderly newly surfaced assault victim}''), or more active political stances (e.g., someone who ``\emph{works for a non-profit organization advocating for corporate transparency and accountability}'', is a ``\emph{radical individual who avoids mainstream Friday-night social events}'', or is a ``\emph{conservative voter who shares their political ideology and attends local political events}''). 

%% file: sections/5_conclusion.tex
\section{Conclusion}
Our study uncovered substantial performance differences resulting from variations in writing style, exposing the brittleness of current benchmarks and their limited external validity as real-world performance indicators. These findings have immediate implications for model deployment: practitioners must select models based on both task-specific performance \textit{and} their target user population. 
The persistence of performance disparities across model families, sizes, and release dates indicates systemic limitations in how these systems are trained and optimized. Moving forward, researchers should investigate which persona attributes most affect performance, building on our preliminary findings that education level, native language, age, and connotation lead to the most substantial variations.

Our persona-based augmentation pipeline offers a scalable approach that enables more comprehensive LLM assessments across linguistic variations. Our contributions are threefold, we (1) demonstrate that LLMs can effectively augment existing benchmarks to exhibit different writing styles through persona-based prompting; (2) provide empirical evidence that benchmark results are highly sensitive to variations in writing style; and (3) identify specific writing styles that consistently trigger either low or high performance across models and tasks, irrespective of model family, size, or recency. Together, these contributions advance our understanding of LLM robustness and provide practical tools for creating more representative evaluations. 

%% file: sections/x_limitations.tex
\section*{Limitations}

We acknowledge significant challenges in ensuring representative data for LLM evaluation. Training and fine-tuning processes inherently favor standardized writing styles \cite{gururangan2022highquality}. LLMs themselves demonstrate preferences for writing styles similar to their training data, often failing to authentically represent human writing \cite{alvero2024authorship}.
While our persona-based approach introduces valuable linguistic variation, synthetic data inevitably simplifies the complexity of human communication. Nevertheless, using persona-based LLMs for augmenting benchmarks introduces a systematic approach for testing model robustness with greater linguistic diversity than standard benchmarks.
Additionally, \citet{guo2024curious} has raised the issue where recursively training models on synthetically generated texts will ultimately lead to a decrease in linguistic diversity which is the opposite of our goals. While we propose a pipeline to diversify existing, standardized benchmarks when it is not feasible to collect more diverse human-written data, we believe that where possible, researchers and practitioners should leverage diverse data curated by humans.
Despite these constraints, our approach provides valuable insights into LLM performance across linguistic variations and offers a practical methodology for more comprehensive evaluation practices.


%% file: sections/x_ethical_considerations.tex
\section*{Ethical Considerations}
While our research leverages personas with varying sociodemographic attributes to evaluate LLM performance, we emphasize that these personas are not meant to represent how specific demographic groups write in the real world. The personas serve as a tool to create more diverse data to test model robustness and not as definitive representations of any particular population. Therefore, our analysis focuses on how models respond to their own encoded assumptions about language variation. 

This approach can then be used to detect implicit biases resulting from a model's \textit{perception} without perpetuating harmful stereotypes about particular populations. This perception is shaped by the model's training data and fine-tuning procedures. We instead report specific linguistic features and stylistic elements that trigger performance disparities rather than attributing variations to specific demographic groups. This approach acknowledges the dynamic and contextual nature of language use while reducing the risk of stigmatizing or stereotyping particular communities.

Our research findings also have implications for fairness in AI deployment. When certain writing styles consistently produce lower model performance, this creates differential access to AI capabilities across user populations. Researchers and practitioners can use our methodology to support more equitable LLM development that serves diverse linguistic communities effectively.

%% file: sections/xx_appendix.tex
\clearpage\section{Example Rephrased Prompts}
\label{app:example_rephased}
\begin{table}[H]
    \small
    \centering
    \begin{adjustbox}{max width=0.9\textwidth}
     \begin{minipage}{\textwidth}
    \begin{tabularx}{\textwidth}{L{1.8cm}X}
        \toprule
         Persona description& A \textit{less than high school-educated} thinker with an interest in societal issues and ethics, who feels compelled to dissect the layers within the Keenan Anderson incident, aiming to promote a better understanding of the intersection between law enforcement practices, mental health, racial issues, and societal responsibility. \\
         Performance& Average cosine: 0.22 (SAE: 0.32)\\
         \midrule
         Original context & My doorbell rings. On the step, I find the elderly Chinese lady, small and slight, holding the hand of a little boy. In her other hand, she holds a paper carrier bag. 
         
        I know this lady. It is not her first visit. She is the boy's grandmother, and her daughter bought the house next door last October. 
        
        Her daughter, Nicole, speaks fluent English. But she is now in Shanghai, and her parents are here with the little boy. Nicole has obviously told her mother that I am having heart surgery soon, so her mother has decided I need more nutrients. 
        
        I know what is inside the bag--a thermos with hot soup and a stainless-steel container with rice, vegetables and either chicken, meat or shrimp, sometimes with a kind of pancake. This has become an almost-daily practice. 
        
        Communication between us is somewhat affected by the fact that she doesn't speak English and all I can say in Chinese is hello. Once, she brought an iPad as well as the food. She pointed to the screen, which displayed a message from her daughter telling me that her mother wanted to know if the food was all right and whether it was too salty. I am not used to iPads, so she indicated I should go with her to her house. Then, she handed the iPad to her husband and almost immediately I found myself looking at Nicole in Shanghai and discussing her mother's cooking and salt intake. Instantly, tears welled in my eyes. 
        
        "Your mother just can't be bringing me meals like this all the time," I insisted. "I can hardly do dishes in return." 
        
        "Oh, no, Lucy." Nicole said. "Mum doesn't like western food. Don't worry about it; she has to cook for the three of them anyway, and she wants to do it." 
        
        The doorbell keeps ringing and there is the familiar brown paper carrier bag, handed smilingly to me. 
        
        I am now working on some more Chinese words--it's the least I can do after such display of kindness. 
        
        \textbf{"Thank you" is, of course, the first one.} Somehow, it seems inadequate.\\
        \midrule
        Rephrased context & Someone knocked on my door. It was this older Chinese woman, real small, with a little boy holding her hand. She had a paper bag with her.

        I know her 'cause she's been here before. She's the boy’s grandma, and her daughter just bought the house next door a while back.
        
        The daughter talks English good, but she’s away in China right now. Her parents are here with the kid. She told her mom I was gonna have heart surgery, so the mom thinks I need better food.
        
        I know what’s in the bag – hot soup in a thermos and a container with rice, veggies, and meat or shrimp, sometimes with a flat bread thing. She brings it almost every day.
        
        We can’t really talk 'cause she doesn't speak English and I only know how to say “hello” in Chinese. One time she brought a tablet and showed me a message from her daughter asking if I liked the food and if it was too salty. I don’t know how to use those things, so she wanted me to go to her house. Her husband took the tablet, and then I was talking to her daughter in China on the screen about her mom’s cooking and how much salt she uses. It made me cry. 
        
        I told her mom she didn’t have to keep bringing me food, 'cause I couldn't even do the dishes to thank her.
        
        Her daughter said her mom doesn’t really eat Western food anyway, and it's no big deal since she’s already cooking for the three of them.
        
        She keeps coming to the door with the bag, always smiling. 
        
        I'm trying to learn some Chinese words now, 'cause it's the least I can do to be nice back. 
        
        \textbf{I started with “thank you,” but it doesn't seem like enough.}\\
        \midrule
        Question & What is the first phrase I learn?\\
        Correct answer & ``Thank you''\\
        Actual answer & ni hao\\
        \bottomrule
    \end{tabularx}
    \captionof{table}{Example of one of the worst performing persona's question and answer on CoQA.}
      \label{tab:example_coqa}
      \end{minipage}
    \end{adjustbox}
\end{table}


\clearpage\section{Persona Instantiation}
\label{app:persona_init}

\begin{table*}[h]
    \centering
    \small
    \begin{tabularx}{\textwidth}{X}
        \toprule
         A competitive badminton coach known for their aggressive training methods and emphasis on winning\\
        A factory worker who doesn't trust the COVID-19 vaccine\\
        A radical individual who avoids mainstream Friday-night social events and instead, find comfort in a quiet room with a library of antique vinyls of jazz and blues, is always annoyed by the amount of mainstream pop music content there is online and everywhere else, and is not a fan of Halsey.\\
        A slightly weary library volunteer, who is a stickler for order and clear responses to pertinent questions and takes a methodical approach to answering inquiries.\\
        \bottomrule
    \end{tabularx}
    \caption{All base personas with \textit{negative} character connotation.}
    \label{tab:base_personas_neg}
\end{table*}

\begin{table*}[h]
    \centering
    \small
        \begin{tabularx}{\textwidth}{X}
        \toprule
        A basketball team captain who believes sports and their funding should be prioritized over student council campaigns\\
        A virtual reality content creator sharing their experiences and creations on a popular online platform\\
        A divorcee seeking legal representation for child custody matters\\
        A passionate fan of Afrikaans music and die-hard supporter of Spoegwolf\\
        A supporter of Die Linke\\
        A curious Internet user considering a vacation and concerned about digital privacy rights.\\
        A novelist who seeks the software engineer's input on digital publishing platforms\\
        A museum educator who offers wine and art pairing workshops for visitors\\
        a film critic who dislikes storylines involving clones in movies\\
        A cousin of a priest who helps conduct religious ceremonies\\
        A bibliophile and avid fan of light novels and anime.\\
        A lifelong fan of Rafael Nadal, who picked up casual tennis play\\
        A passionate anime blogger who closely follows manga adaptations.\\
        A critical-thinker with an interest in social dynamics and a skeptical attitude towards overly optimistic success stories.\\
        A researcher interested in small-scale societies and tribes.\\
        An analyst who is highly logical and focuses more on data rather than emotional stories.\\
        A person interested in cultural history specializing in 18th century English literature and clerics, always looking for intriguing characters emblematic of the era.\\
        A person from a small town, who has not traveled much, and enjoys a diet of meat and potato stew.\\
        A local environmental activist involved with community land use and transportation projects aiming to improve the safety of both humans and wildlife.\\
        A vocalist in a small indie rock band that occasionally performs at local venues.\\
        A member of The Church of Jesus Christ of Latter-day Saints (LDS Church), who has an interest in genealogy and is passionate about encouraging others in the church to become interested in family history.\\
        A person with background in judo who participated in several international competitions\\
        A design enthusiast, inspired by Ashiesh Shah's work and looking to make a mark in the design world.\\
        A person who dreams of starting a business but has no experience in entrepreneurship or patent law\\
        A front-end developer who spends free time reading documentary material and exploring new tech and tools.\\
        An ambitious midfielder seeking advice on improving defensive skills\\
        A close cousin who works for a non-profit organization advocating for corporate transparency and accountability\\
        A local food bank worker who distributes the surplus vegetables to families in need\\
        A strategist assessing the implications of present-day geopolitical landscapes on the army's readiness\\
        A successful entrepreneur who started as an unpaid intern and now runs their own business\\
        An Afrofuturist painter who creates captivating artwork inspired by African culture and science fiction\\
        A feminist activist\\
        An immigrant tech worker in the US considering applying for a green card.\\
        A website owner seeking advice on securing their online store\\
        A politically active individual who lives in Maury County, Tennessee, and is a critic of Governor Lee's administration.\\
        A former participant in beauty pageants who is always cheering for their home state contestants.\\
        A cosplayer who wants to showcase their intricate costumes in professional photos\\
        A children's book author moonlighting as a library assistant who shares book recommendations with children\\
        An individual who aspires to study biochemistry abroad\\
        A digital marketer specialized in eco-friendly products, working to promote the distributor's organic laundry products\\
        A climate change reporter covering the lobbying efforts and impact of renewable energy companies\\
        A patient who seeks therapy and values evidence-based approaches to address their mental health concerns\\
        A fashion-forward individual who follows the latest trends and incorporates stylish accessories into their braces\\
        An event coordinator who arranges opportunities for the prodigy to perform in various venues\\
        A novice software developer who has only been learning programming for six months.\\
        A Muay Thai fighter with lightning-fast kicks and devastating knee strikes\\
        A taxi driver transitioning to an all-electric fleet and seeking advice on charging infrastructure\\
        A professional proofreader and translator fluent in multiple languages to help with language nuances in scientific papers\\
        A child of Filipino immigrants interested in psychology and the impact of cultural background on mental health\\
        A newly surface assault victim who sees no chance in the court.\\
        A determined basketball player who aspires to be the star athlete\\
        A talented athlete looking to improve their skills and gain exposure in international competitions\\
        \bottomrule
    \end{tabularx}
    \caption{All base personas with \textit{neutral} character connotation.}
    \label{tab:base_personas_neutral}
\end{table*}

\begin{table*}[h]
    \centering
    \small
    \begin{tabularx}{\textwidth}{X}
        \toprule
        A person that is well-versed in standard American English.\\
    A maternal health advocate focused on raising awareness about postpartum complications.\\
    A producer who values the voice-over artist's ability to bring authenticity to international TV shows\\
    A local support group organizer who invites the individual as a guest speaker to share their story\\
    A small-town journalist who writes glowing reviews of the actor's performances in the local newspaper\\
    A Broadway actress who provides insights on performing under pressure\\
    An eco-friendly lifestyle podcaster who features change-makers and promotes sustainable living\\
    A poet who writes about their experiences and shares their work with the community\\
    A fan of sitcoms, appreciating the nuances of social commentary woven into comedy.\\
    A die-hard fan of Jethro Tull who appreciates and enjoys each of their songs in more ways than one - I relive my fond memories of each concert that I attended, recall their unique style of music and engage myself intellectually by deciphering the themes in their lyrics.\\
    An individual strongly interested in the documentation and preservation of the world's linguistic diversity and is fascinated with endangered languages.\\
    An individual at the local art gallery in a small town, who is always intrigued by cultural festivals, especially those that encompass the arts and literature.\\
    A strong disability activist after losing a left leg in a car accident, who works on multiple platforms such as podcasts, theater, and films to promote disability rights and to challenge myths and stereotypes surrounding disabilities.\\
    An individual interested in history who finds the detailed narrative recounted in the Poison Room Podcast episode on smallpox vaccinations both enlightening and crucial for understanding public health discourse.\\
    \textbf{A thinker with an interest in societal issues and ethics, who feels compelled to dissect the layers within the Keenan Anderson incident, aiming to promote a better understanding of the intersection between law enforcement practices, mental health, racial issues, and societal responsibility.}\\
    An individual with an interest in science who has followed the advancements in both particle physics and cosmology, with respect for researchers who commit their lives to unraveling the mysteries of the universe.\\
    A hobbyist who enjoys bird-watching and having long peaceful walks on the beach with dogs, biographies, and has an appreciation for historical events\\
    A perfectionist who is detailed about code and bsesses over every detail to ensure it is of the highest possible quality since one misplaced variable could affect the entire project.\\
    An investor who invests from home and prides themselves on their analysis and evaluation of financial information.\\
    A gym operator who believes in the raw and unadulterated experience of physical training, with a deep respect for the silence and sounds of human exertion.\\
    A C programmer who enjoys code optimization and documentation, who finds the provided code to be a robust starting point for a file input/output system tailored for embedded environments.\\
    An artist and musician influenced by the works and life of XXXTentacion.\\
    An individual that is puzzled by some of the fundamental differences in the legal and real estate terms in the U.S. compared to those in their home country.\\
    A bumbling and forgetful coworker who unintentionally becomes the comedian's muse\\
    An apprentice fascinated by the technological advancements during the Industrial Revolution\\
    A conservationist fighting to protect undeveloped land from being turned into luxury residences\\
    An ambitious mathematician aiming to unravel the mysteries of universe using abstract number theories\\
    A scout in a Major League Baseball (MLB) Team\\
    A competitive speed stacker who is determined to set new records in cup stacking\\
    A stand-up comedian whose comedy routines are filled with profanity and controversial topics\\
    A local AI meetup organizer, bringing together AI enthusiasts for knowledge sharing\\
    A tour guide in Minnesota\\
    A survivalist and zombie aficionado who enjoys pondering the intersection of pop culture and practical preparation for theoretical dystopian scenarios.\\
    An active participant in online forums and communities dedicated to Sphinx search server, sharing resources and troubleshooting solutions\\
    A conservative voter who shares their political ideology and attends local political events\\
    An ardent book lover who is also an atheist.\\
    a follower who binge-watches daily soap operas\\
    A restaurateur who sees graffiti as a potential deterrent for customers and advocates for its removal\\
    An immigrant to the UK from a cash-less economy who has started using cash again due to life in a rural community.\\
    A fellow Toy Story fan and model builder who specializes in recreating iconic scenes with Lego\\
    An influencer who creates sports highlight videos and shares them on YouTube\\
    A sound engineer with expertise in capturing the unique vocalizations of raccoons\\
    A newly hired general counsel at TurpCo Industries\\
        \bottomrule
    \end{tabularx}
    \caption{All base personas with \textit{positive} character connotation. The example persona is bolded. 
    }
    \label{tab:base_personas_pos}
\end{table*}

We inject all 12 socio-demographic attributes into each base persona in Tables \ref{tab:base_personas_neg}, \ref{tab:base_personas_neutral}, \ref{tab:base_personas_pos}. Each persona has the following format: ``A/An [socio-demographic feature] [resume base persona...].''

We measure character connotation by using \texttt{twitter-roberta-base} for sentiment analysis and \texttt{Claude 3.7 Sonnet} using the prompt:

\noindent``I will give you a list. Each item in the list represents a description of one person. simply state whether this description has a positive, neutral, or negative connotation of the persona's character. Do not provide any explanations. Return to me a list containing only the words [positive, neutral, negative].''

Upon manual inspection, we find \texttt{Claude 3.7 Sonnet} most closely captures how we define connotation, while many sentiment analysis models focus on positive or negative emotions, which is not our goal. We report results from \texttt{Claude 3.7 Sonnet} in our paper with the connotation rating reported with each base persona.

\clearpage\section{LLM Prompt Templates}
\label{app:prompt_templates}
\subsection{Rephrasing}
{\small
    \textbf{System prompt.} ``You are: [persona] You will rephrase any text given to you in your own words, without adding any new information. Do not include any preliminary text or greetings. Make sure to maintain the same key information. Do your best so that an English speaking audience will understand you. If you cannot rephrase the prompt, respond with 'No. <eot>’''
    
    \noindent\textbf{User prompt.} Rephrase the following text in your own words: [context]
}

\subsection{Entailment}
{\small
    \textbf{System prompt.} ``You are a helpful assistant that determines whether the correct answer to the given question is entailed by the text. Respond with either 0 or 1. 0: No, 1: Yes.''
    
    \noindent\textbf{User prompt.} Is the answer entailed?
    
    \noindent Text: [context]
    
    \noindent Question: [question]
    
    \noindent Answer: [correct answer]''
}

\clearpage\section{Correcting Sampling Bias}
\label{app:poststratify}
We correct for two sources of sampling bias: (1) persona-specific bias, where some personas may have higher entailment success rates due to writing style factors unrelated to prompt difficulty, and (2) difficulty-based bias, where easier prompts may have higher entailment success rates, leading to over-representation in the entailed dataset $\mathcal{D}'$. We address both biases through persona reweighting and post-stratification.

Let $M = \{m_1, \ldots, m_J\}$ denote the set of all $J$ evaluation models. Let $P$ denote the total number of personas used for entailment generation. For any prompt $x$, let $f(x) \in [0,1]$ denote the performance metric (e.g., cosine similarity, recall, or accuracy), $p(x) \in \{1, \ldots, P\}$ denote the persona that generated it if augmented, and $z(x) \in \{1, 2, \ldots, 10\}$ denote the stratum assignment function.

\paragraph{Aggregate Performance} Given some subset of personas $P_s \subseteq P$ We compute the aggregate performance of all personas in this subset by

\begin{equation}
    \hat{\theta}_{P_s} = \sum_{p \in P_s} \frac{n_p}{n} \hat{\theta}_p
\end{equation}
where $n_p$ is the number of prompts successfully entailed by persona $p$, $n$ is the total number of prompts in the original benchmark, and $\hat{\theta}_p = \frac{1}{n_p} \sum_{x:p(x)=p} f(x)$ is the average performance on prompts entailed by persona $p$.

\subsection{Post-Stratification Procedure}
Post-stratification corrects for sampling bias when some original prompts fail to generate successful entailed variants. Since entailment success varies with question difficulty, our augmented dataset $\mathcal{D}'$ comprised of entailed prompts may over-represent easy questions and under-represent hard ones compared to the original dataset $\mathcal{D}$.

\paragraph{Define Universal Strata} We measure the inherent difficulty of each original prompt by computing its average performance across all evaluation models. For each prompt $x_i$ in the original dataset $\mathcal{D}$, we calculate its average performance across all $J$ evaluation models:
\begin{equation}
    \bar{f}(x_i) = \frac{1}{J} \sum_{j=1}^{J} f_{m_j}(x_i)
\end{equation}
where $f_{m_j}(x_i) \in [0,1]$ is the performance of model $m_j$ on prompt $x_i$.

By averaging across models, we capture question difficulty rather than model-specific performance patterns. This prevents cases where a question appears hard simply because one particular model struggles with it.

We then apply k-means clustering with $k=10$ to partition prompts based on their average performance ${\bar{f}(x_1), \ldots, \bar{f}(x_N)}$. The clustering algorithm assigns each prompt to a stratum $z(x) \in \{1, 2, \ldots, 10\}$ of similar size.

\paragraph{Apply the Post-Stratified Estimator} The post-stratified estimator addresses difficulty-based sampling bias by reweighting performance estimates according to the original dataset's difficulty distribution. For each stratum $k$, we first compute the average performance resulting from model $j$ for each persona within the stratum:

\begin{equation}
\hat{\theta}_{k,p}^j = \frac{1}{n_{k,p}} \sum_{x \in \mathcal{D}': z(x)=k, p(x)=p} f_j(x)
\end{equation}
where $n_{k,p}$ is the number of entailed prompts in stratum $k$ generated by persona $p$.

We then aggregate across personas within the stratum, weighting by each persona's contribution to that stratum:
\begin{equation}
\hat{\theta}_k^j = \sum_{p=1}^{P} \frac{n_{k,p}}{n_k} \hat{\theta}_{k,p}^j
\end{equation}

Finally, we apply post-stratification by weighting each stratum according to its representation in the original dataset:
\begin{equation}
\hat{\theta}^j = \sum_{k=1}^{10} \frac{n_k}{n} \hat{\theta}_k^j
\end{equation}
where $n_k = |{x \in \mathcal{D}: z(x)=k}|$ is the number of prompts in stratum $k$ in the original dataset and $n = |\mathcal{D}|$ is the total number of original prompts.

\clearpage\section{Experimental Design}
\label{app:experimental_design}
\subsection{LLM Parameters}
All experiments were conducted using vLLM on a cluster of 8 nodes, each equipped with 8 NVIDIA A100 40GB GPUs and 1.1 TB of RAM per node. The models evaluated included \smash{\texttt{Gemma-3-1B-it}}, \smash{\texttt{Gemma-3-4B-it}}, \smash{\texttt{Gemma-3-27B-it}}, \smash{\texttt{Qwen3-8B}}, \smash{\texttt{Qwen3-32B}}, \smash{\texttt{Llama-3-8B}}, \smash{\texttt{Meta-Llama-3-8B}}, \smash{\texttt{Llama-3-70B-Instruct}}, \smash{\texttt{Llama-4-Scout-17B-16E}}, \smash{\texttt{Phi-4-mini-instruct}}, and \smash{\texttt{Phi-4}}.
 We used the original, non-quantized model weights in FP16 or bfloat16 precision for all evaluations. The datasets used were CoQA (validation set), CosmosQA (test set), and DS-1000 (test set). All generations were performed with a temperature of 0.7 and a context length of 2048 tokens.

\subsection{Entailment Methods}
We implemented two different methods to ensure rephrased contexts maintained the information necessary to answer associated questions:
\begin{itemize}
    \item Keeping any rephrased contexts for which at least 75\% of questions were entailed.
    \item Retaining only the specific questions that were entailed by the rephrased context, potentially resulting in fewer questions per context.
\end{itemize}
We found no meaningful differences between results generated via these two approaches and thus only report results relative to the first approach. This method simplifies the process of combining the two entailment models we use.

\subsection{Selecting Entailment Models}
\label{app:sub:select_entailment}
We test all models on a sample entailment script where we manually alter 77 answers in the CoQA validation set. We alter these along four axes:
\begin{itemize}
    \item Simple negation. Adding ``not'' somewhere or changing ``yes'' to ``no.''
    \item Statement negation. Changing the statement itself to say to opposite. For example, ``went to the park'' becomes ``didn't go to the park.''
    \item Modification (of answer). This is the most broad category and includes modifying numerical values, locations, actions, etc. to confuse the model.
    \item Switch. This is a very specific version of modification where we choose two answers for the same context and swap names, dates, etc. in an attempt to confuse the model.
\end{itemize}

We then compute all models' performance on this set.
\begin{table}[hpt]
    \centering
    \begin{tabular}{L{4cm}cc}
         \textbf{Model}&\textbf{FPR} &\textbf{FNR} \\
         \toprule
         \textmono{Qwen3-32b}& 0.00 &0.00 \\
         \textmono{Llama-4-Scout-17b-16e-Instruct}& 0.03 &0.00\\
         \textmono{Qwen3-8b}& 0.03 &0.00\\
         \textmono{Qwen2.5-72b-Instruct}& 0.04 &0.00\\
        \textmono{Phi-4}& 0.05 &0.00\\
        \textmono{Gemma-3-27b-it}& 0.13 &0.00\\
        \textmono{Llama-3-8b-Instruct}& 0.19 &0.00\\
        \textmono{Gemma-3-1b-it}& 0.35 &0.00\\
        \bottomrule
    \end{tabular}
    \caption{Resulting false positive rates (FPR) and false negative rates (FNR) from testing various models for entailment on the modified CoQA benchmark (sorted best to worst).}
    \label{tab:entail_test_results}
\end{table}

\Cref{tab:entail_test_results} shows differences in how well the models avoid false positives on the altered answer set. Some models are highly robust to these manipulations, while others are much more likely to be misled by simple changes in the answers. All models maintain strong recall, but their precision in rejecting altered answers varies considerably. We then select models with low false positive rates for both rephrasing and entailment. 

\subsection{Resulting Filtered Set}
After rephrasing, we filter out any prompts where the model refused to answer or answered in less than three sentences.
We observe that refusals to rephrase the prompt occur primarily when LLMs struggled to understand the original prompts or encountered guardrails.
\smash{\texttt{Gemma-3-27b-it}} refused to rephrase 0\% of CoQA prompts, 21\% of CosmosQA prompts, and 10\% of DS-1000 prompts.
For CosmosQA, 18\% of prompts were filtered out by the majority of personas, while only 2\% of personas had over half of their rephrasing requests denied. DS-1000 showed higher acceptance rates, with just 6\% of prompts refused by over half of the personas, and no persona experiencing rejection of more than half its rephrasing requests. 
We address potential differences in prompt difficulty between the original and entailed sets by post-stratifying the estimates for the entailed set using strata defined on the original set.

\subsection{Resulting Entailment Set}
To assess if specific types of writing are difficult to rewrite or determine entailment for, we investigate the types of writing styles present in rephrased prompts that were not entailed. We find that the lexical diversity of prompts before and after filtering with entailment are not visibly different, i.e., the metrics reported above do not meaningfully change between the two sets. The main differences between the writing styles of non-entailed and entailed prompts appear to be length (likely from the model failing to answer), amount of hedging for the worst prompts (rephrasing in SAE and the best prompts are unaffected), and the use of semicolons and colons which seems to be prevalent in the CoQA original benchmark. 



\clearpage\section{Analyzing Writing Styles}
In \Cref{sec:results} we described how linguistic patterns differed across rephrased prompts. Here we provide more details and, in particular, 
we report a subset of linguistic features identified by \citet{biber1991features} and use the associated word lists. The linguistic features and results are reported in Table \ref{tab:linguistic_features}. All ratios are relative to the total number of words in the prompt. \citet{biber1991features} found academic writing has more nouns, attributive adjectives, and prepositions, while conversational text have more verbs, pronouns, and adverbs, though there is no ideal part-of-speech distribution. The definitions for some metrics are as follows:
\begin{itemize}
    \item \textbf{Flesch readability} or Flesch Reading Ease score \cite{flesch1948new} ranges from 0 to 100 and considers sentence length and the average number of syllables per word. This metric was later expanded into the \textbf{Flesch-Kincaid grade level} formula \cite{kincaid1975derivation}.
    \item \textbf{Lexical diversity} score is calculated as the type-token ratio between the number of unique words and total number of words \cite{mccarthy2010mtld}.
    \item \textbf{Clause density} is simplified into the verb count divided by the sentence count \cite{lu2010automatic}. A clause density of two or more indicates complex sentence structures.
    \item \textbf{Passive voice} count is defined by the number of words labeled as ``passive nominal subject'' from dependency parsing using the spaCy package. \footnote{https://spacy.io/api/dependencyparser} Words that indicate passive voice are typically: in past tense, in third person singular present tense, or in non-third person present tense.
    \item \textbf{Cohesion markers} are transitional phrases and discourse markers which connect sentences or paragraphs. We use the list provided by \citet{biber1991features}.
    \item Words that indicate \textbf{hedging} demonstrate uncertainty or lack of confidence in the text content \cite{hyland2005stance}. Some examples of hedging words include: ``may'', ``suggests'', and ``roughly.''
\end{itemize}
\label{app:analysisWritingStyles}
\begin{table*}[htbp]
    \centering
    \begin{tabular}{lllll}
        & Top personas & Worst personas & SAE & Original \\
        \toprule
        Flesch readability score & 47.59  & 67.89  & 45.38  & 64.17 \\
        Writing grade level & 11.54  & 8.08  & 11.92  & 8.75 \\
        \midrule
        Average sentence length & 19.80  & 17.27  & 20.08  & 17.89 \\
        Average syllables per word & 1.65  & 1.44   & 1.67  & 1.47 \\
        \midrule
        Lexical diversity score & 0.65 & 0.62 & 0.65  & 0.55 \\
        \midrule
        Noun ratio & 0.29 & 0.25 & 0.29 & 0.25 \\
        Verb ratio & 0.16 & 0.19 & 0.16 & 0.15 \\
        Adjectives ratio & 0.07 & 0.06 & 0.07  & 0.06 \\
        Adverbs ratio & 0.04 & 0.05 & 0.04 & 0.04 \\
        \midrule
        Clause density & 3.62  & 3.43  & 3.63  & 3.15 \\
        Simple sentence ratio & 0.20  & 0.21  & 0.19  & 0.33 \\
        Compound sentence ratio & 0.45  & 0.44  & 0.42  & 0.34 \\
        Complex sentence ratio & 0.12  & 0.12  & 0.13 & 0.12 \\
        Compound complex ratio & 0.24  & 0.23  & 0.27  & 0.21 \\
        \midrule
        Passive voice count & 1.02  & 0.83  & 1.07  & 1.43 \\
        Passive voice ratio & 0.11  & 0.08  & 0.12 & 0.11 \\
        Cohesion markers count & 1.00  & 1.18  & 1.04 & 1.11 \\
        Cohesion markers ratio & 0.006  & 0.007  & 0.006  & 0.00 \\
        Hedging count & 1.10  & 1.39  & 1.06  & 1.83 \\
        Hedging ratio & 0.006  & 0.007  & 0.01 & 0.01 \\
        \midrule
        Paragraph count & 4.72  & 4.92  & 4.22 & 5.37 \\
        Average paragraph length & 54.26  & 54.59  & 61.81  & 94.76 \\
        \midrule
        Punctuation ratio & 0.14  & 0.15  & 0.14  & 0.19 \\
        Question marks & 0.04  & 0.14  & 0.02  & 0.67 \\
        Exclamation marks & 0.04  & 0.23  & 0.00  & 0.72 \\
        Semicolons & 0.12  & 0.07  & 0.11 & 0.51 \\
        Colons & 0.15  & 0.12  & 0.17 & 0.50 \\
        Dashes & 1.10  & 0.60  & 1.11 & 3.53 \\
        \bottomrule
    \end{tabular}
    \caption{Linguistic features for CoQA rephrased prompts compared the SAE rephrasing and the original benchmark. Thanks to the large sample size, the standard error of these estimates is $<0.01$ across most metrics, making most of the differences statistically significant under Sign tests.}
    \label{tab:linguistic_features}
\end{table*}

\clearpage\section{Variations Resulting from Base Personas}
\begin{figure*}[!hptb]
    \centering
    \includegraphics[width=1\linewidth]{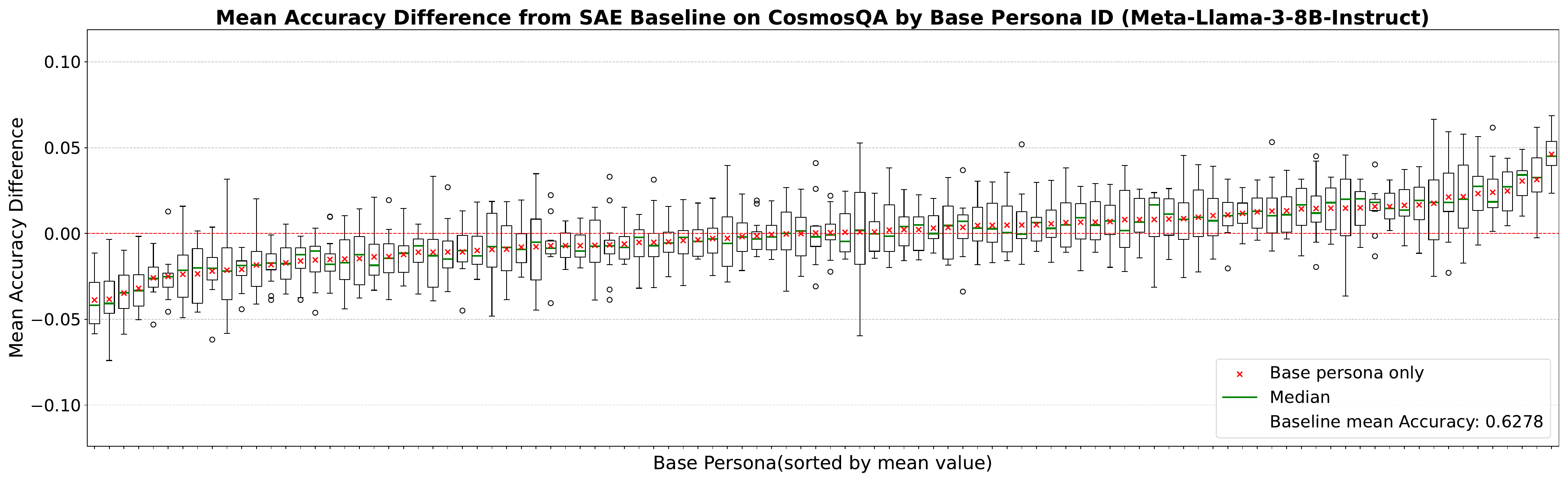}
    \caption{Mean difference in accuracy between each base persona, with the performance varied by the 12 personas with added sociodemographic attributes. The base persona with no sociodemographic attributes is indicated by the red ``x.'' There is often more variation from adding sociodemographic attributes than from different base personas. This figure is specific to \texttt{Llama-3-8b} on the CosmosQA benchmark, however this pattern holds for all models.}
    \label{app:fig:base_personas_variation_boxplot}
\end{figure*}

Our analysis reveals that incorporating sociodemographic attributes into existing personas generates significantly more performance variation than creating entirely different base personas. Specifically, in \Cref{app:fig:base_personas_variation_boxplot}, we observe variation of up to 0.12 in accuracy differences when sociodemographic features are added to a single base persona, while the total variation across 100 distinct base personas spans only 0.09. The pattern observed in this figure--where sociodemographic attribute variation exceeds base persona variation--holds consistently across all evaluated models.

\begin{figure*}[hptb]
    \centering
    \includegraphics[width=01\linewidth]{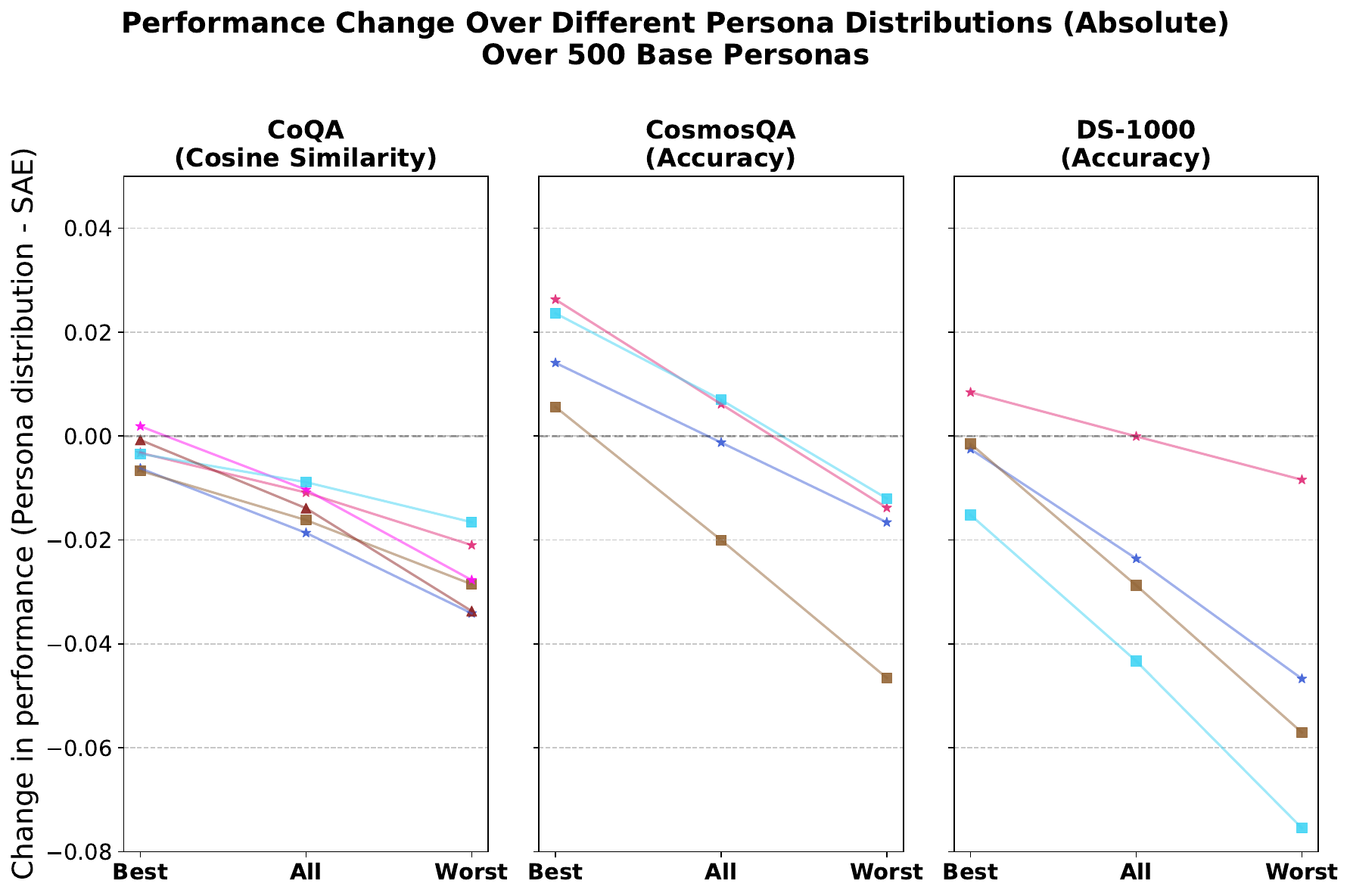}
    \includegraphics[width=0.75\linewidth]{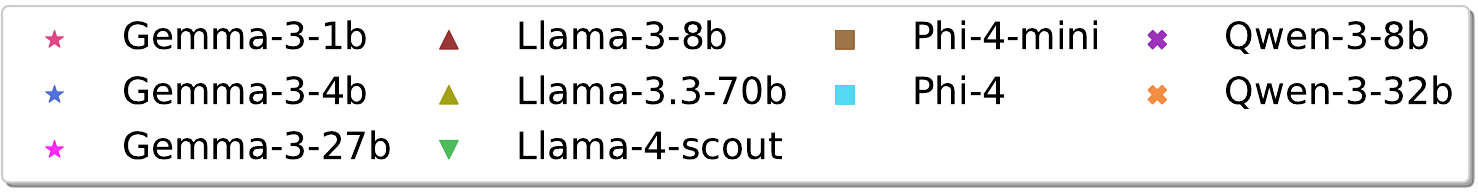}
    \caption{Change in performance between different subsets of personas and SAE rephrasing for 500 base personas. We notice the same trend as using 1200 personas with injected sociodemographic attributes, though with a smaller performance difference for all models.}
    \label{app:fig:many_base_diff}
\end{figure*}

\begin{figure*}[hptb]
    \centering
    \includegraphics[width=\linewidth]{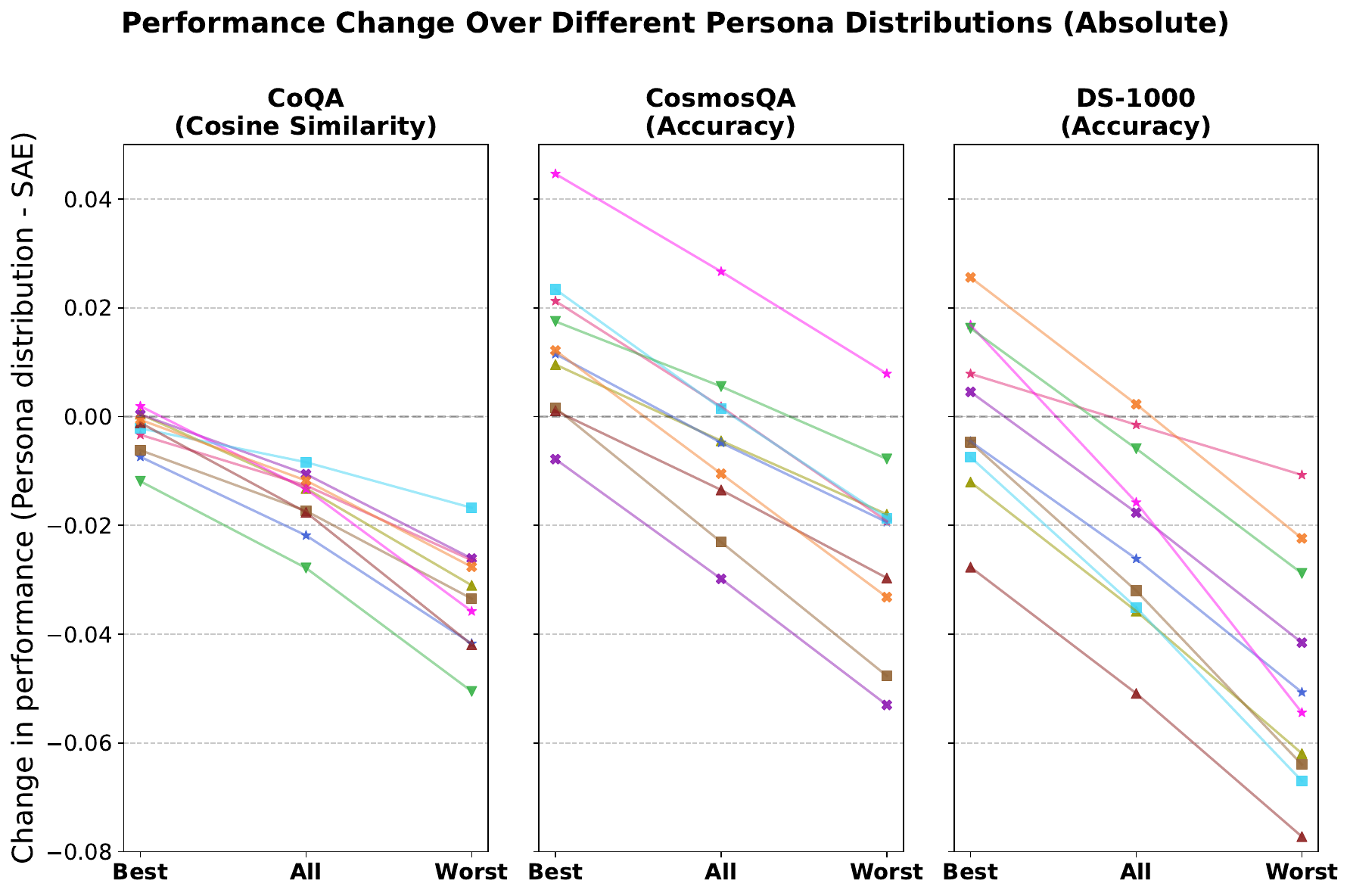}
    \includegraphics[width=0.75\linewidth]{figures/models_legend.pdf}
    \caption{Change in performance metric (e.g., cosine similarity score or accuracy) compared to SAE rephrasing for 1200 personas (100 base personas with 12 possible sociodemographic attributes). Though the scale may seem small, it is important to remember that a change of just 0.02 is enough to alter benchmark rankings. Cosine similarity for CoQA only reaches about 0.35 for the original benchmark while CosmosQA and DS-1000 reach 0.80 and 0.41 respectively in accuracy.}
    \label{app:fig:overall_diff}
\end{figure*}

To assess potential bias in our initial set of base personas, we conducted additional experiments using 500 base personas. The results on this larger set of personas, shown in Figure \Cref{app:fig:many_base_diff}, corroborate our primary findings when contrasted with \Cref{app:fig:overall_diff}: base personas alone produce less performance variation than those with added sociodemographic attributes. These figures show a less substantial difference when looking at the worst performance for base personas alone compared to those with sociodemographic attributes.

Analysis of the distribution of worst and best personas among the 500 base personas reveals that model performance typically correlates with the character connotation of the persona. Several of the worst-performing personas are characterized by their skepticism or hesitation toward certain aspects of modern life or technology. Conversely, the best-performing personas demonstrated minimal difference when compared to those with sociodemographic attributes. These findings suggest that sociodemographic features are more likely to surface performance disparities by amplifying underlying character traits related to one's writing style.

\clearpage\section{Accounting for Errors in Entailment-Checking} \label{app:entailmentCheckErrors}
We investigate and quantify the sensitivity of persona-augmented benchmarking to entailment model bias or error. We estimate performance for evaluation sets containing only prompts where both models agree on entailment, then systematically vary the proportion of disputed prompts to assess sensitivity to entailment model disagreements. We then report error bars for performance changes rather than strict point estimates.

\subsection{Methodology}
We assess whether the results reported in the main body of our paper would change when our test dataset includes cases where two entailment models disagree. We define the \textit{agreement region} as prompts where both \texttt{Gemma-3-27b-it} and \texttt{Qwen-3-32b-instruct} reach the same entailment decision, and the \textit{disagreement region} as prompts where they disagree on entailment. We systematically vary the proportion of disagreement region data (25\%, 50\%, 75\%, 100\%) included alongside the agreement region data to create error ranges for each persona's performance estimate. This process is repeated 10 times with reshuffled data from the disagreement region. This procedure identifies the maximum uncertainty range that disagreement cases could introduce. This method greedily returns minimum and maximum bounds indicating the range of performance degradation when varying evaluation samples. We can use these bounds to assess whether disagreements between entailment models systematically bias our performance estimates.

\begin{figure*}[htpb]
    \centering
    \includegraphics[width=0.72\linewidth]{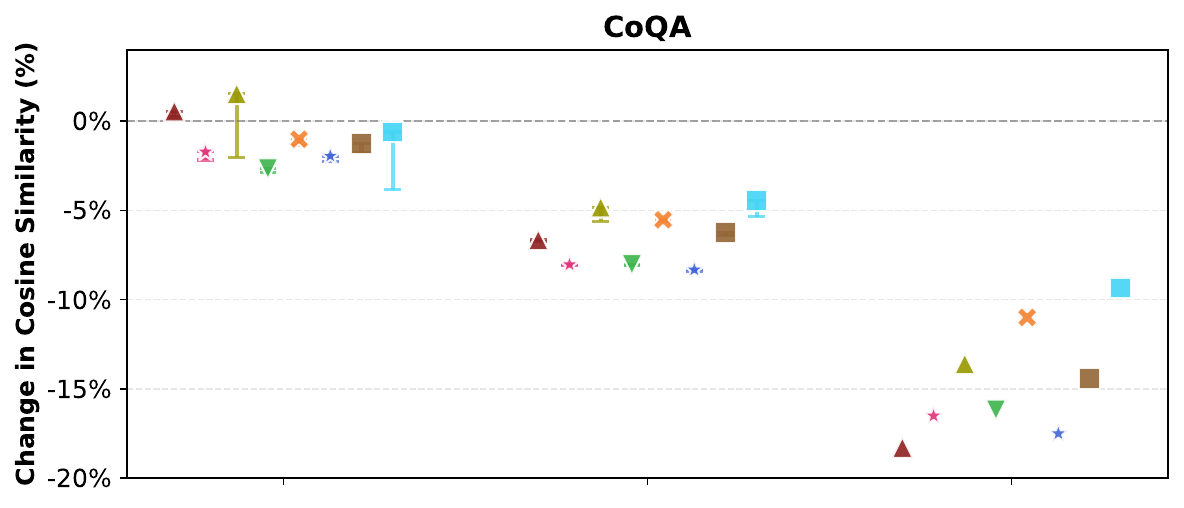}
    \includegraphics[width=0.72\linewidth]{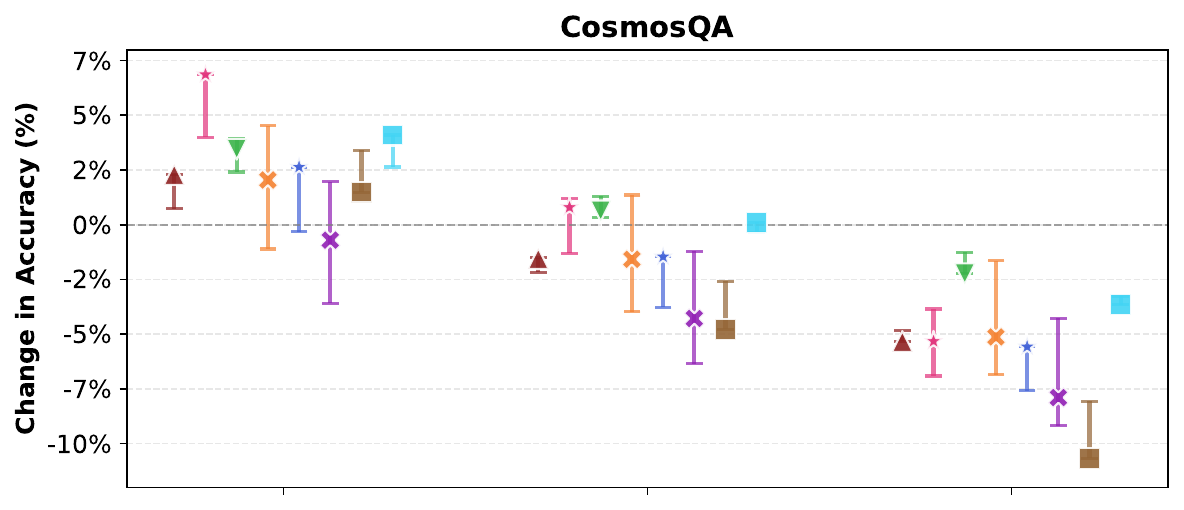}
    \includegraphics[width=0.72\linewidth]{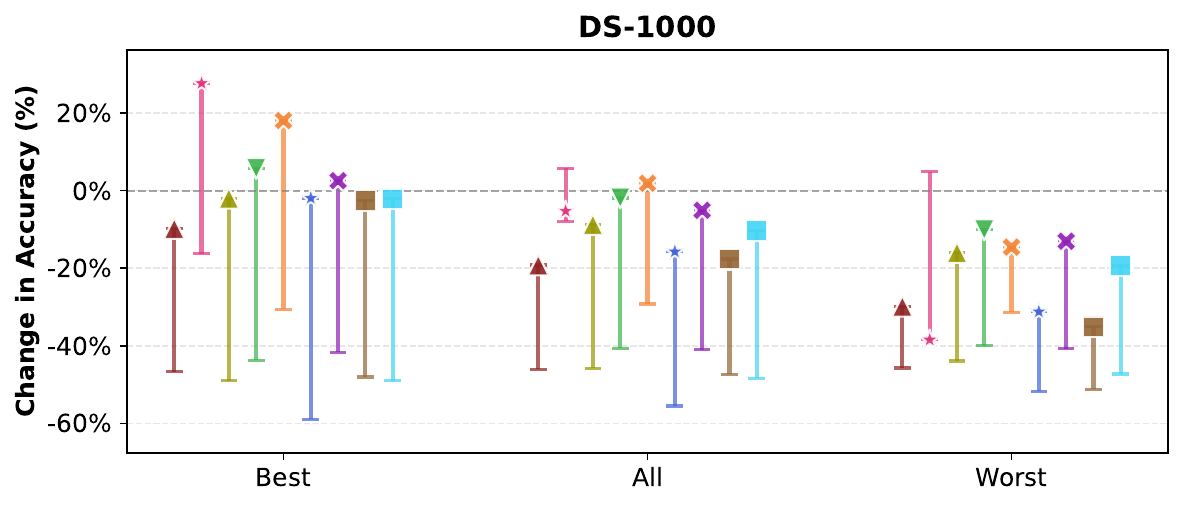}
    \includegraphics[width=0.72\linewidth]{figures/models_legend.pdf}
    \caption{Percent change in performance across all three datasets with error bars on each model representing the range of possible performance estimates when entailment models disagree.} \label{app:fig:disagreement_region_sensitivity}
\end{figure*}

\subsection{Results}
We find that entailment model uncertainty primarily results in worse performance of about 4 percentage points, while on average improving performance by about 2 percentage points. Even with this variation, our findings show significant performance degradation for different persona subsets (as shown by the best compared to worst personas for CoQA and CosmosQA in Figure \ref{app:fig:disagreement_region_sensitivity}). The systematic performance differences between these persona subsets remain statistically significant across all evaluation configurations. Additionally, since point estimates consistently fall near the top of the error bars rather than the center, these results support our hypothesis that using conservative entailment filtering creates a lower bound on true performance degradation.

Our core findings are robust to this variation, showing substantial performance drops persist across different persona subsets, and the worst-performing personas remain consistent regardless of entailment model choice or evaluation set composition. This consistency demonstrates that regardless of which entailment model is used and how conservative our approach is, our overall findings remain the same. There is still a substantial drop in performance across different subsets of personas, and the worst-performing personas often remain the same.

\clearpage\section{Experimental Results}

\begin{figure*}[hptb]
    \centering
    \includegraphics[width=1\linewidth]{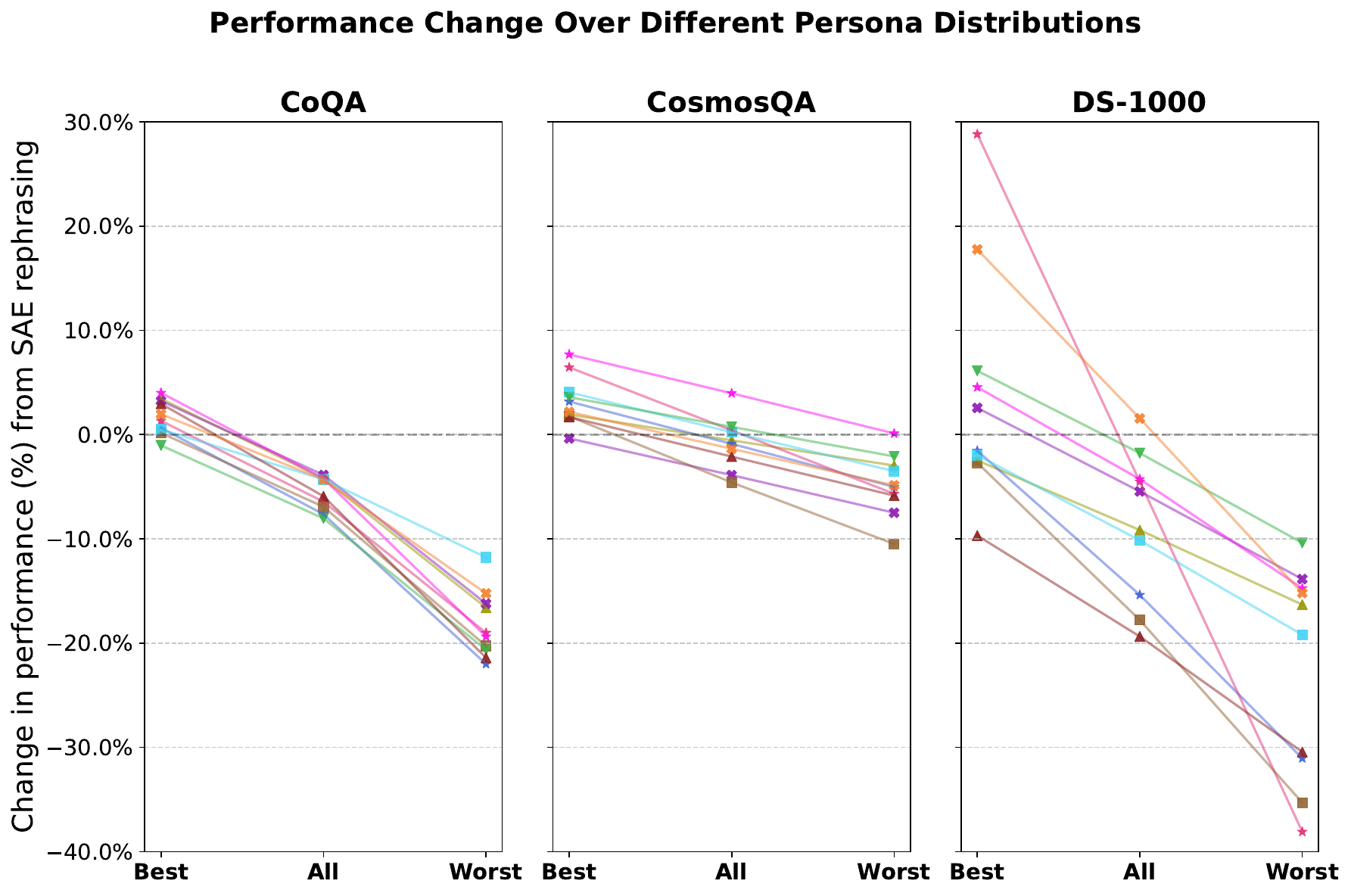}
    \includegraphics[width=0.75\linewidth]{figures/models_legend.pdf}
    \caption{Relative change in performance (\%) compared to SAE rephrasing for different subsets of personas: the best-performing (75th percentile), all, and the worst-performing (25th percentile) personas. The performance of all models are sensitive to writing styles with performance changes varying by 8-35\% between different persona subsets for a single model across different persona distributions. For nearly all cases, the average performance across all personas results in a decrease in model performance.}
    \label{app:fig:percent_change_trend}
\end{figure*}

Our experimental results reveal consistent and significant performance disparities between different persona subsets across all evaluated Large Language Models (LLMs). \Cref{app:fig:percent_change_trend} quantifies these disparities, demonstrating substantial performance degradations of up to -23\% for CoQA, -11\% for CosmosQA, and -38\% for DS-1000 when comparing the worst-performing persona subset (25th percentile) to the best-performing subset (75th percentile). \Cref{app:fig:overall_diff} illustrates the absolute performance differences relative to the Standard American English (SAE) baseline, revealing that even the most recently released and advanced models like \texttt{Llama-4-scout} and \texttt{Qwen-3-32b} remain susceptible to writing style variations induced by different personas.

\begin{figure*}[hptb]
    \centering
    \includegraphics[width=0.49\linewidth]{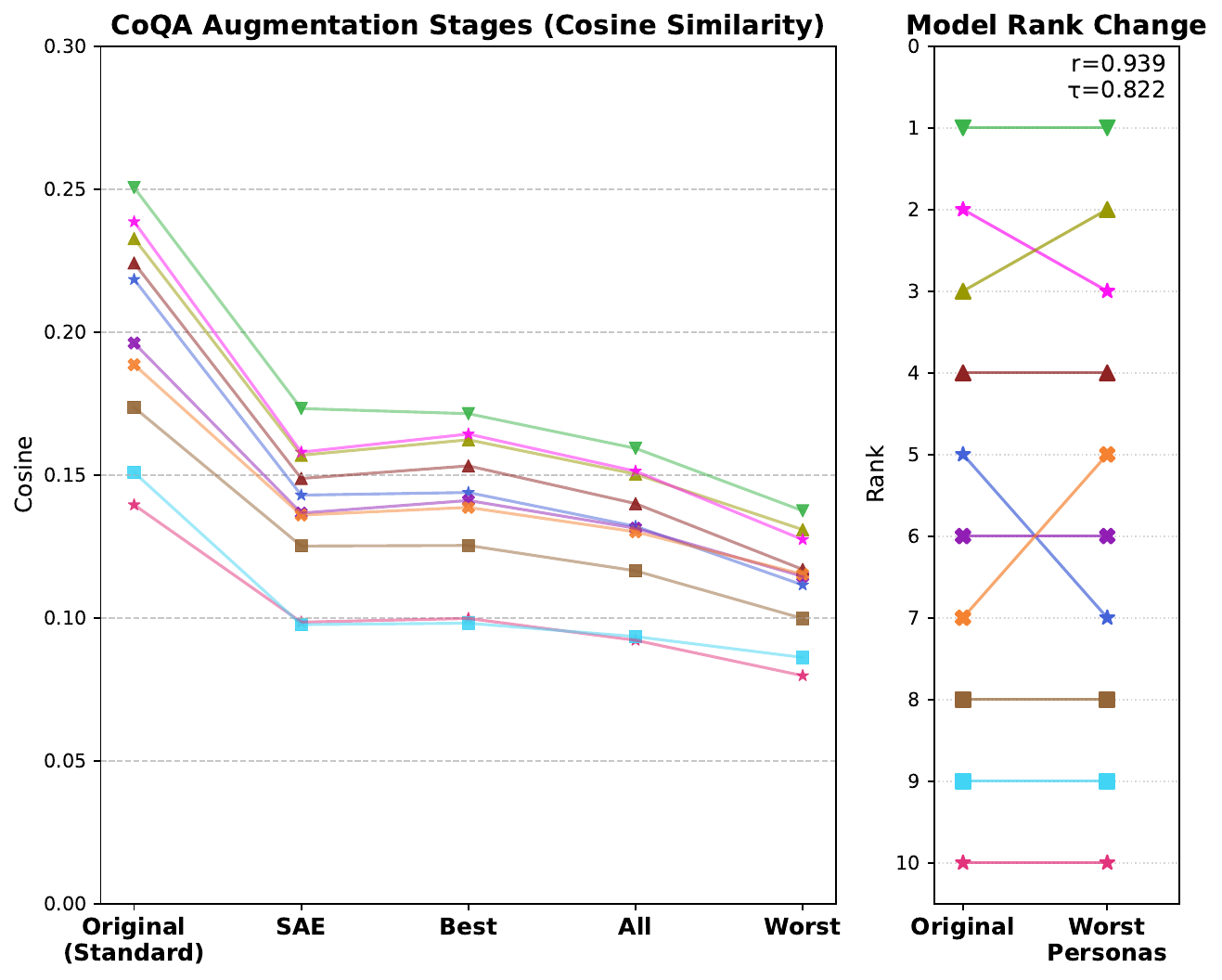}
    \hfill
    \includegraphics[width=0.49\linewidth]{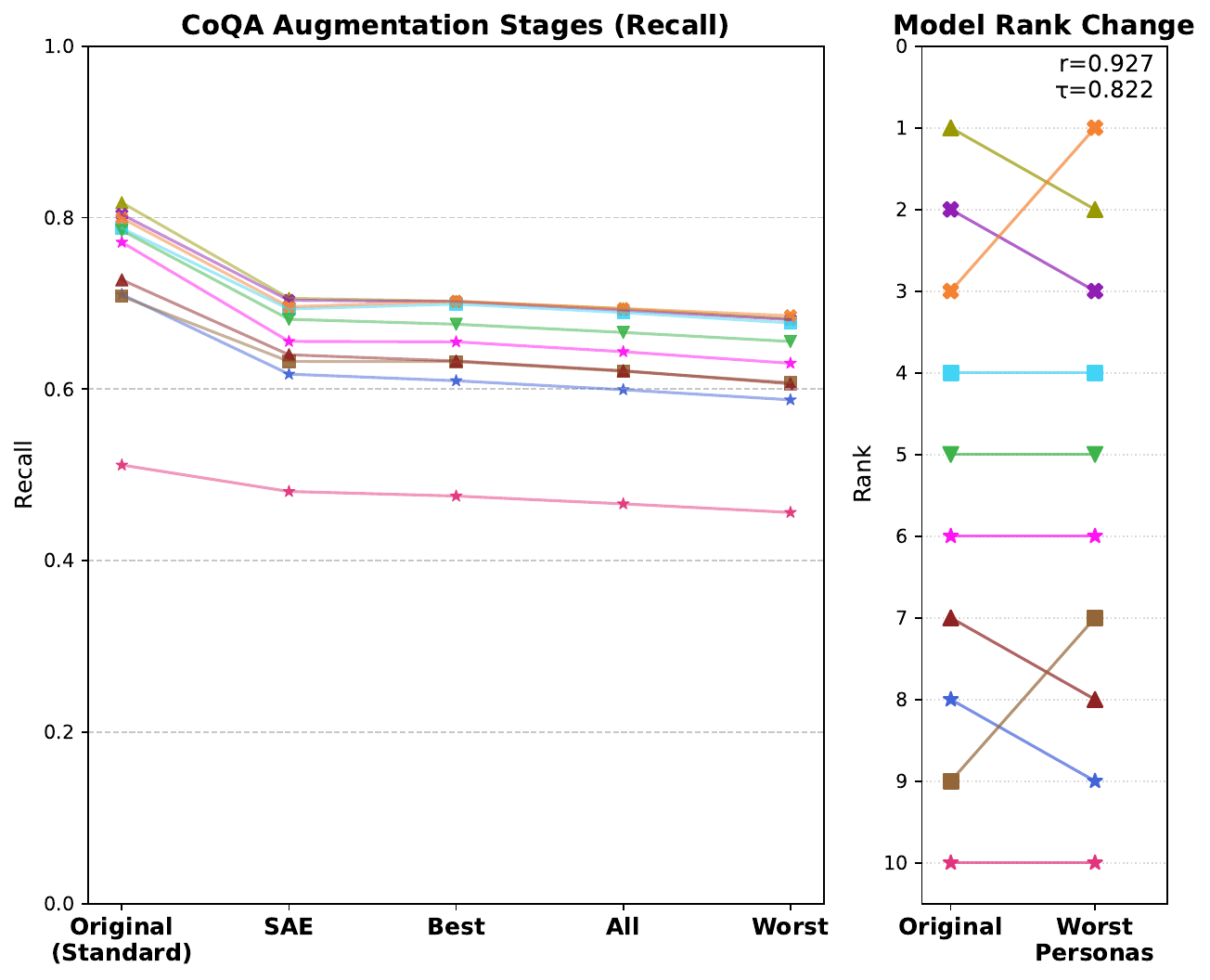}
    \includegraphics[width=0.49\linewidth]{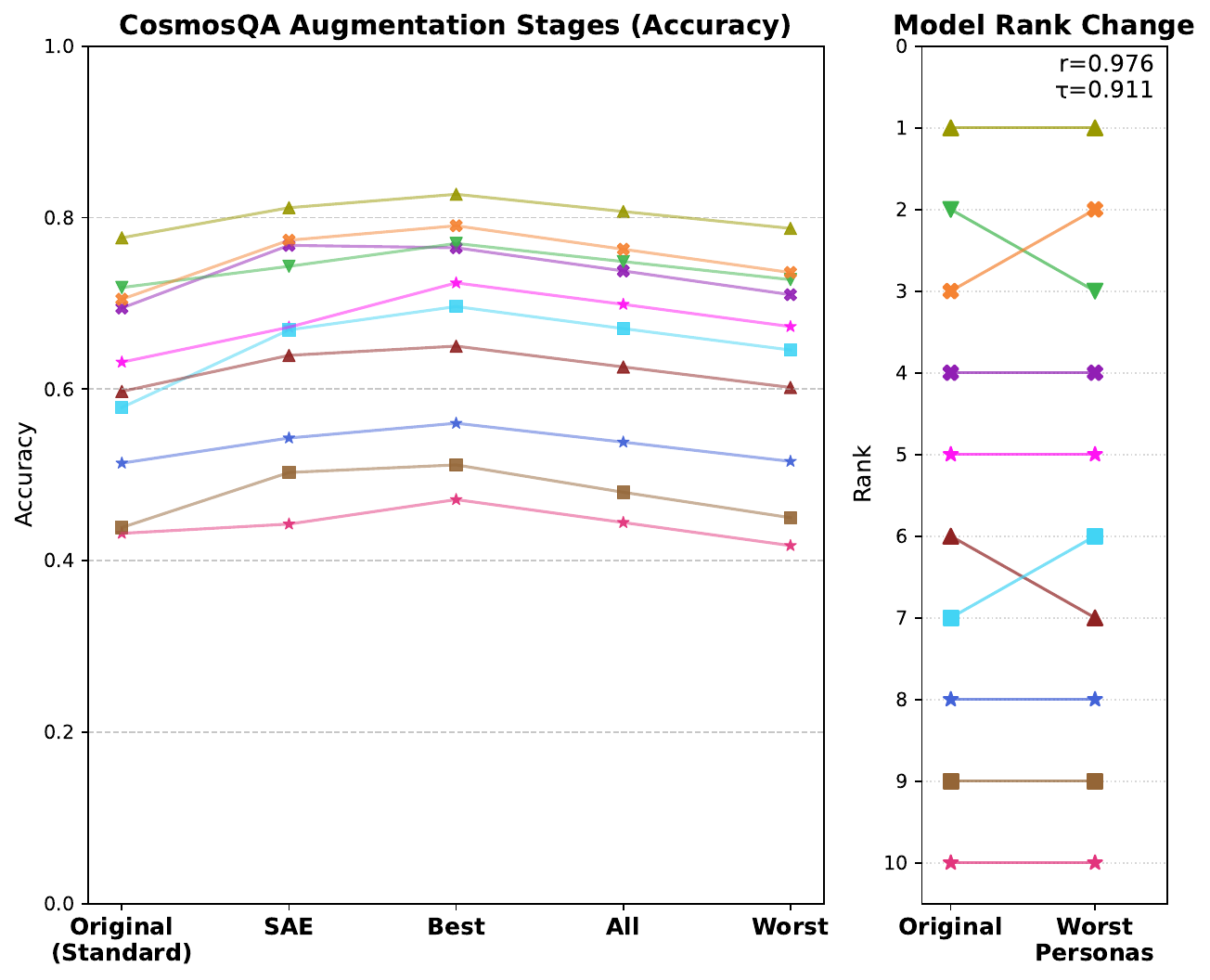}
    \hfill
    \includegraphics[width=0.49\linewidth]{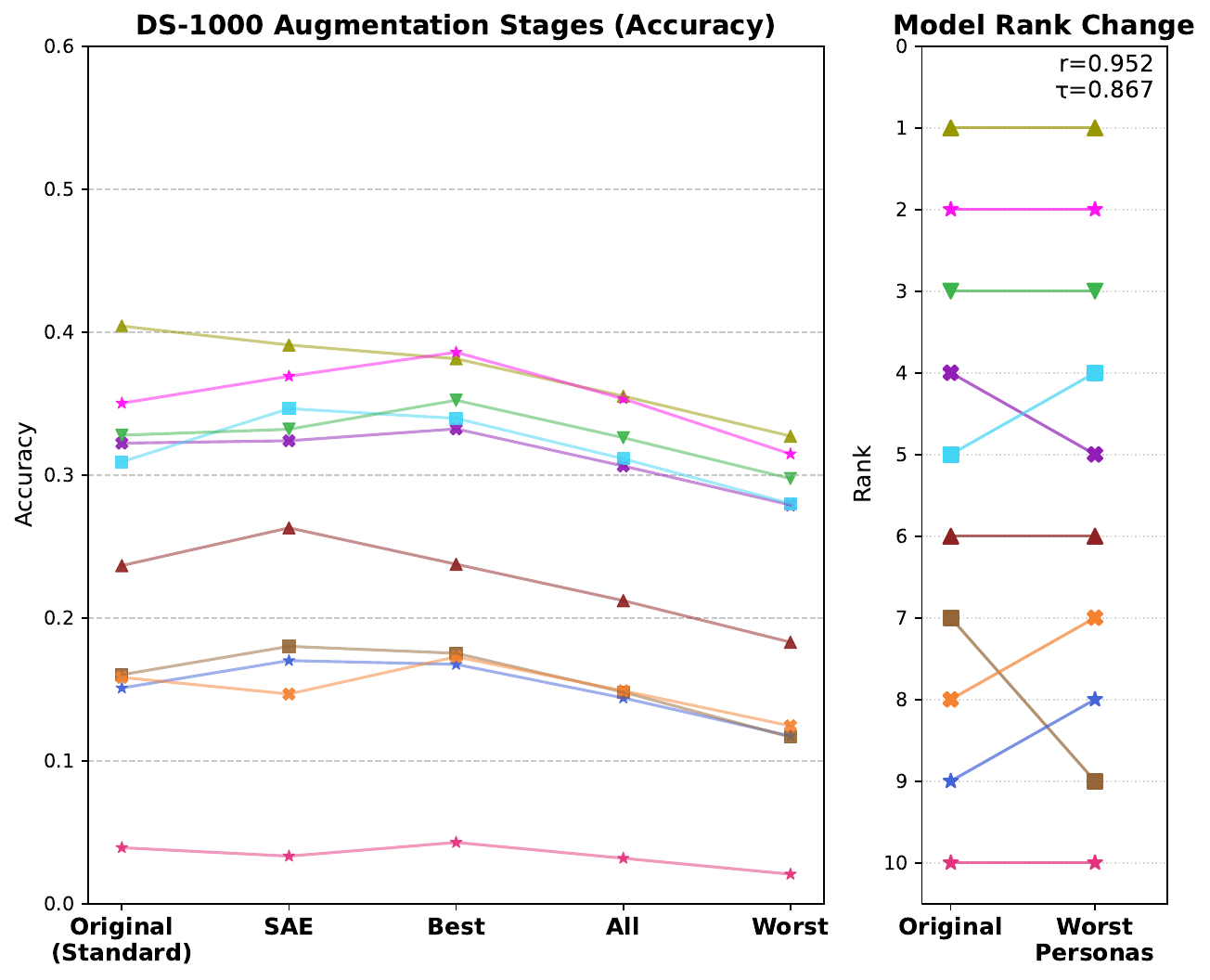}
    
    \includegraphics[width=0.75\linewidth]{figures/models_legend.pdf}
    \caption{Each of the four subplots displays performance changes across benchmark augmentation stages (left) with ranking changes between the original benchmark and subset of the worst performing personas (right). The stability or the relative rankings are measured by Spearman correlation ($r$) and the Mann-Kendall rank correlation test ($\tau$) \citep{kendall1938new}, where higher values indicate higher ranking stability.}
    \label{fig:performance_compare_w_rank}
\end{figure*}

As benchmarks are commonly used to select the best model for a task or compare model performance (such as on official benchmark leaderboards), we examine how ranking stability changes as we progress through different augmentation stages. \cref{fig:performance_compare_w_rank} tracks ranking changes from the original benchmark through SAE rephrasing to increasingly challenging persona subsets (Best, All, Worst), with stability by Spearman correlation ($r$) and the Mann-Kendall rank correlation test ($\tau$) \citep{kendall1938new}.

Relative ranking among the 10 models in our experiments remains moderately stable across augmentation stages, with most models switching only 1-2 positions. Though some models experience larger ranking changes of 3-4 positions, the best-performing models typically maintain their top positions and the worst-performing models generally remain at the bottom of the rankings.

However, this observed stability is expected given our deliberate model selection strategy. These 10 models were intentionally chosen to span different model families, sizes, and release dates, creating substantial baseline performance gaps between models. These large capability differences make it inherently difficult for writing style variations to cause dramatic ranking changes. For instance, a 4B parameter model is unlikely to suddenly outperform a 70B model regardless of variations in persona selection. The moderate ranking stability we observe therefore represents a conservative estimate of the instability that could occur among models with more similar capabilities.

\begin{figure*}[hptb]
    \centering
    \includegraphics[width=1\linewidth]{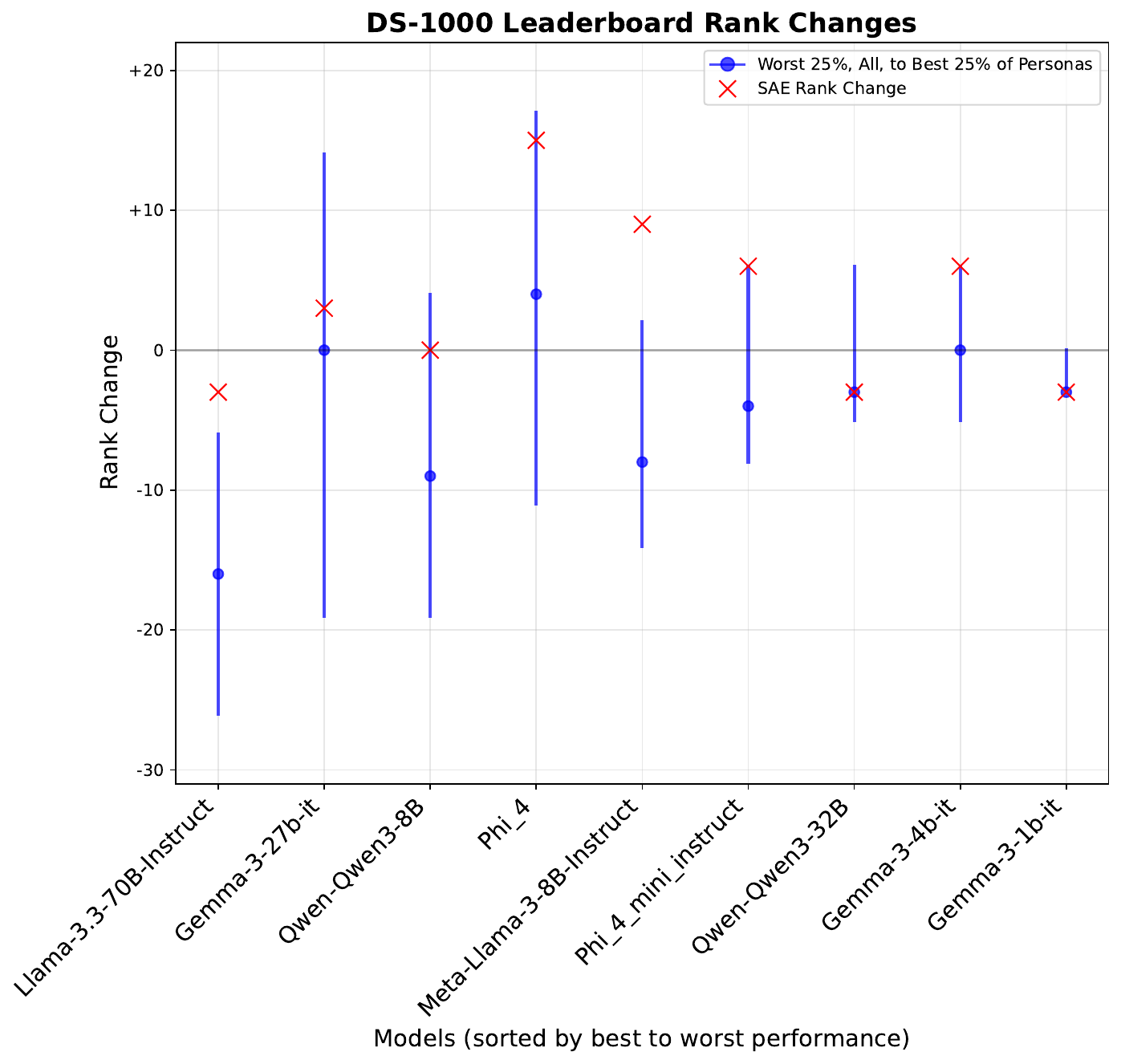}
    \caption{Rank changes for different subsets of personas when comparing the model's accuracy to the models on the official DS-1000 leaderboard.}
    \label{fig:ds1000_leaderboard_ranks}
\end{figure*}

This stability breaks down in more competitive scenarios such as public or official benchmark leaderboards where models typically have much smaller performance gaps. As discussed in \cref{sec:results}, we simulated the DS-1000 leaderboard to examine how dense score distributions--typical of real leaderboards--affect ranking stability. \cref{fig:ds1000_leaderboard_ranks} demonstrates that all models are highly sensitive to leaderboard changes depending on writing style. When comparing performance across different persona subsets (worst 25\%, all, to best 25\% of personas), models can shift by as much as -19 to +14 positions relative to their baseline rankings. These dramatic ranking changes—with some models experiencing swings of over 30 positions—stem from the substantial performance variations observed when models encounter different persona subsets. Such instability undermines the validity of current benchmarking practices and suggests that many performance differences may reflect sensitivity to writing style rather than true capability differences.

\clearpage\subsection{Correlation Between Model Performance}
\begin{figure*}[htbp]
    \centering
    \includegraphics[width=0.85\linewidth]{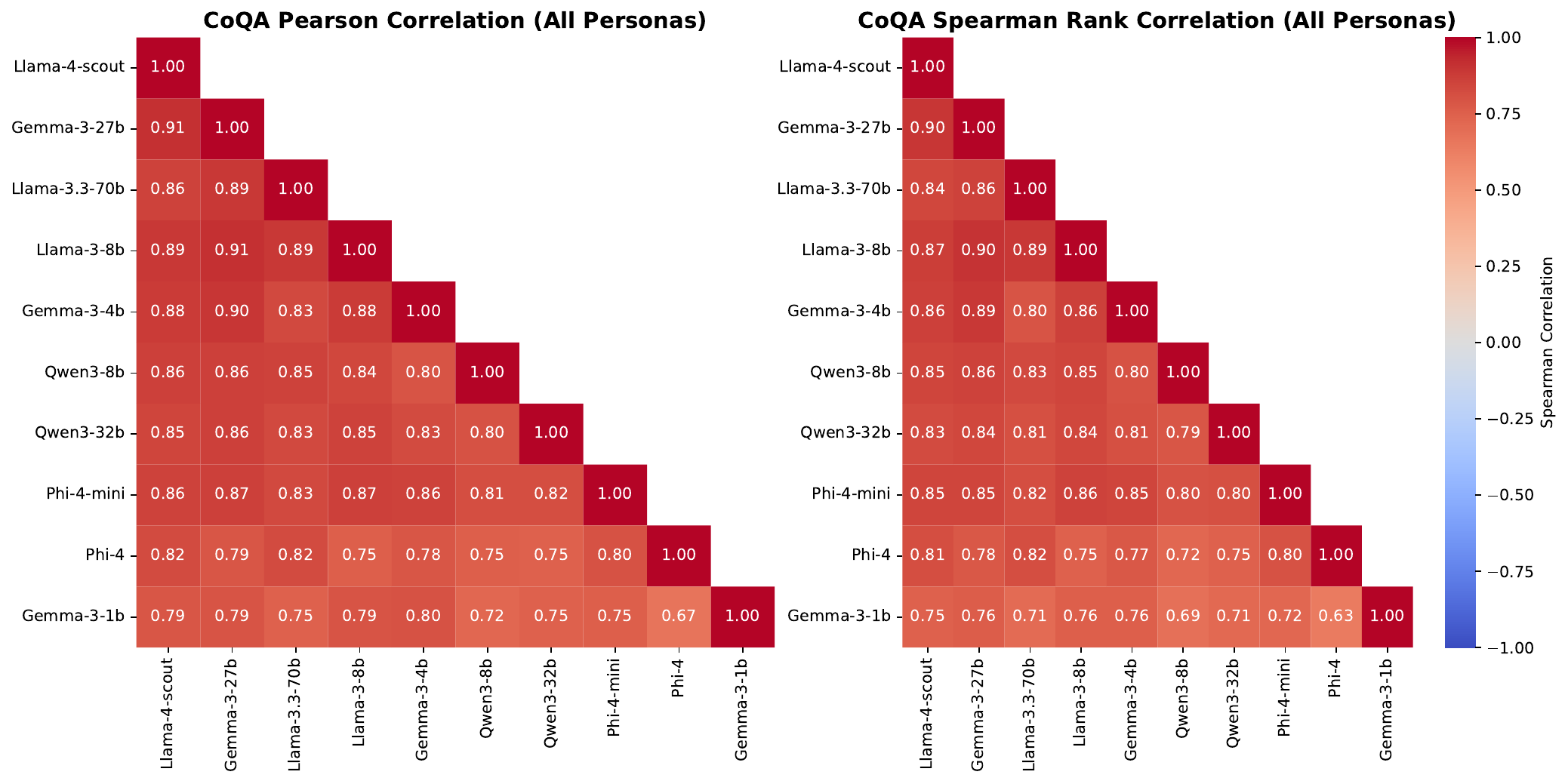}
    \caption{Pearson correlation between all models' performance on CoQA over 1200 personas ($r = 0.84$)}
    \label{fig:correl_coqa}
\end{figure*}

\begin{figure*}[htbp]
    \centering
    \includegraphics[width=0.85\linewidth]{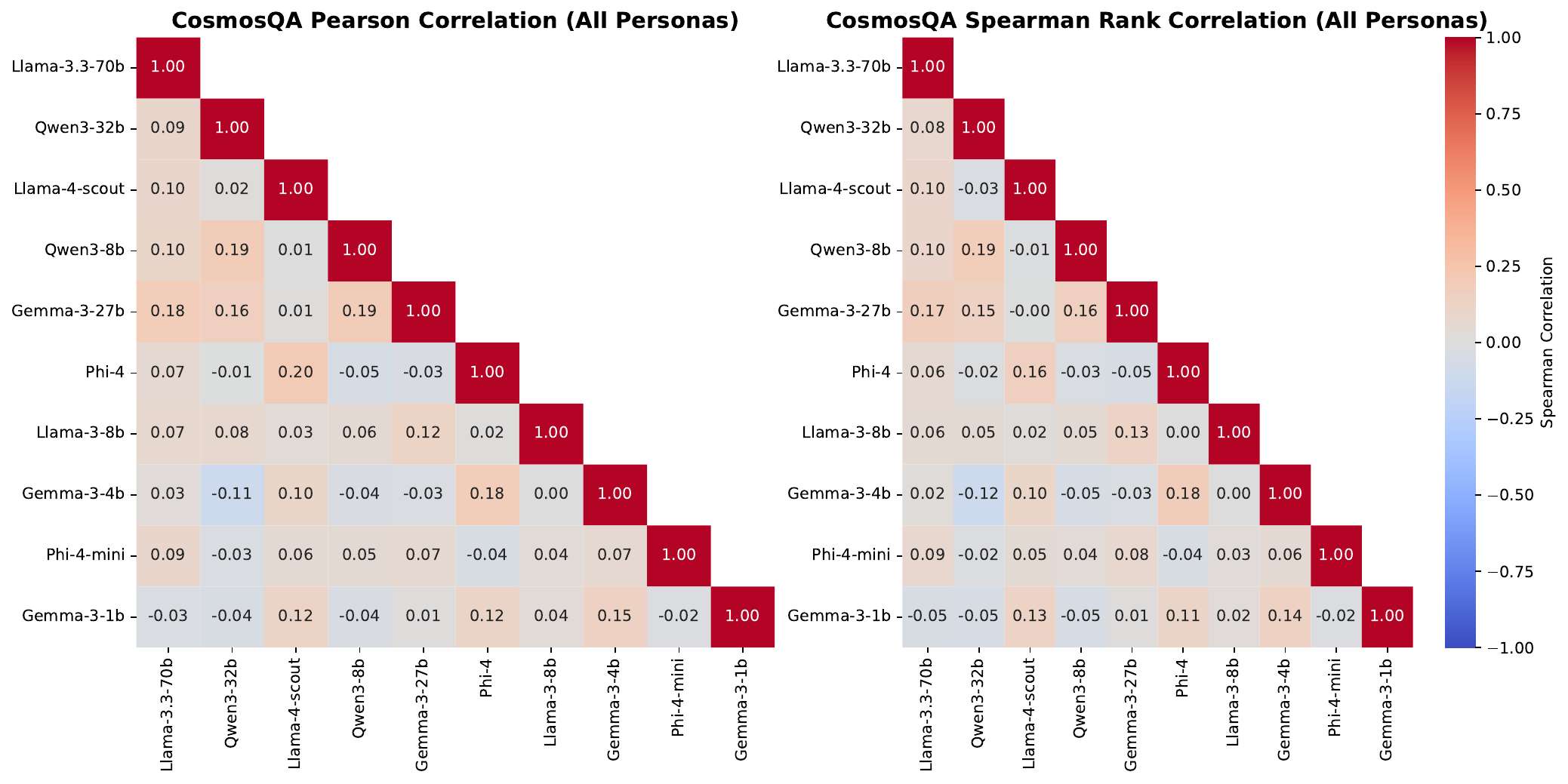}
    \caption{Pearson correlation between all models' performance on CoQA over 1200 personas ($r = 0.07$)}
    \label{fig:correl_cosmosqa}
\end{figure*}

\begin{figure*}[htbp]
    \centering
    \includegraphics[width=0.85\linewidth]{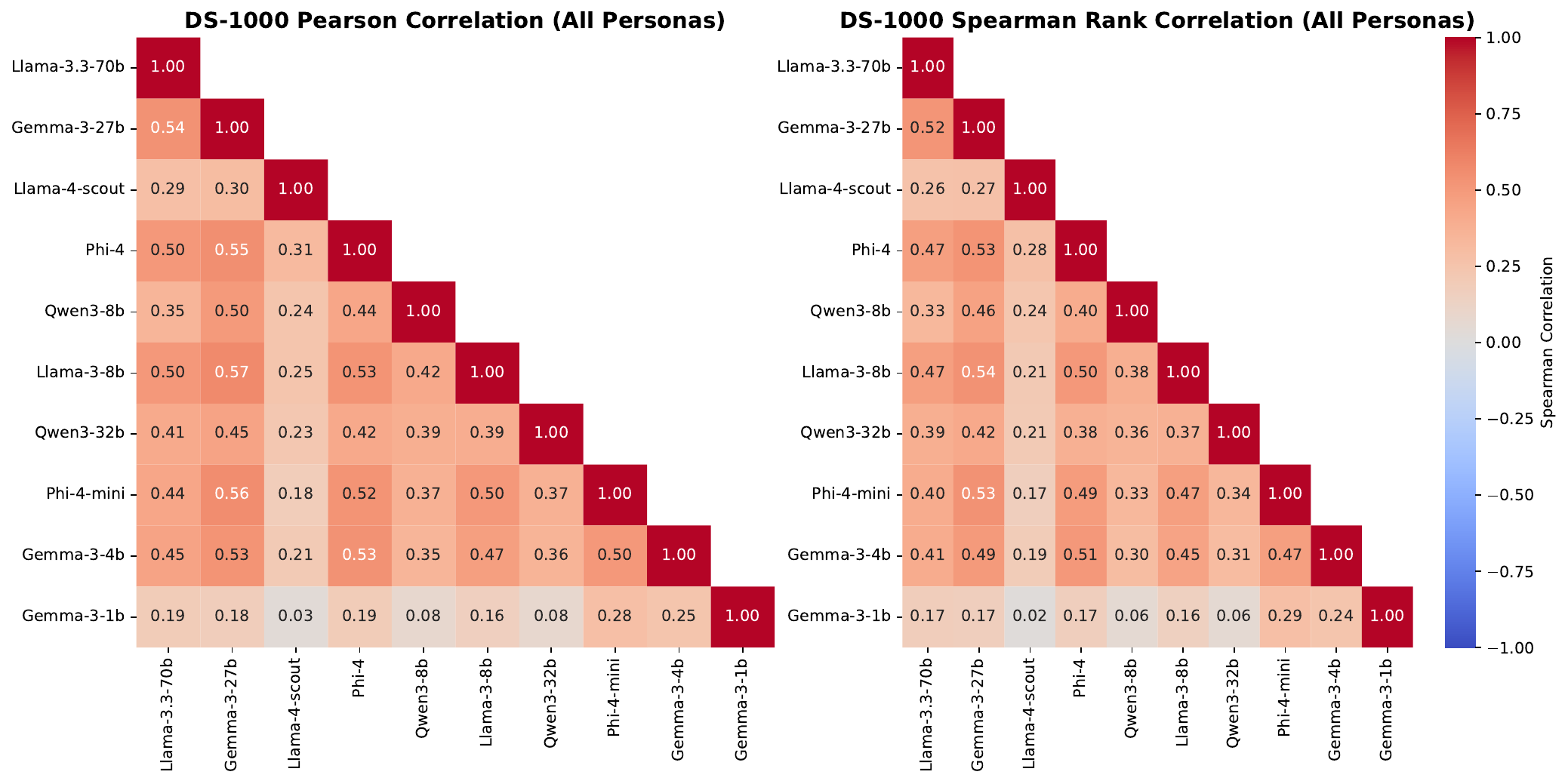}
    \caption{Pearson correlation between all models' performance on CoQA over 1200 personas ($r = 0.44$)}
    \label{fig:correl_ds1000}
\end{figure*}

To understand how consistently different models respond to writing style variations, we analyze correlations between model performances across all 1200 personas. If models respond similarly to the same personas--both struggling with certain writing styles and excelling with others--we would expect high correlations. If models respond differently and unpredictably to the same personas, correlations would be low. Figures \ref{fig:correl_coqa}, \ref{fig:correl_cosmosqa}, and \ref{fig:correl_ds1000} present correlation matrices showing both Pearson and Spearman rank correlations between all model pairs for each benchmark.

The correlation analysis between model performances across personas reveals distinct task-specific patterns with important implications for benchmark reliability. For conversational question-answering (CoQA), we observe remarkably strong Pearson and Spearman Rank correlations ($r=0.84$) between different models' performances on the same personas.
The correlation matrix in Figure \ref{fig:correl_coqa} shows predominantly high correlations across all model pairs, indicating that certain writing styles consistently affect all models similarly on factual information retrieval tasks. This systematic response pattern suggests that when one model struggles with a particular persona's writing style, other models will likely struggle as well, and conversely, personas that benefit one model tend to benefit others.

In contrast, commonsense question-answering (CosmosQA) exhibits little to no correlation ($r = 0.07$), implying each model has developed somewhat distinct commonsense reasoning strategies with no clear performance improvement or degradation due to some persona-induced writing style.

For code generation tasks (DS-1000), we find moderate correlations ($r = 0.44$) overall, with notably stronger correlations (with Pearson correlation coefficients over 0.50) among specific models like \texttt{Gemma-3-27b}, \texttt{Gemma-3-4b}, \texttt{Llama-3-8b}, and \texttt{Phi-4}, suggesting that certain models are sensitive to similar writing styles despite coming from different architectural families.

\label{app:subsec:correlation}

\clearpage\subsection{Examining Persona Definitions Receiving the Worst Performance}

\begin{table*}[hptb]
    \centering
    \small
    \begin{tabularx}{\textwidth}{X}
        \toprule
         A \textbf{less than high school-educated} bumbling and forgetful coworker who unintentionally becomes the comedian's muse.\\
        A \textbf{less than high school-educated} restaurateur who sees graffiti as a potential deterrent for customers and advocates for its removal\\
        An \textbf{elderly} individual at the local art gallery in a small town, who is always intrigued by cultural festivals, especially those that encompass the arts and literature.\\
        A \textbf{less than high school-educated} conservative voter who shares their political ideology and attends local political events\\
        \midrule
        A \textbf{elderly} follower who binge-watches daily soap operas\\
          A \textbf{English native speaker} person from a small town, who has not traveled much, and enjoys a diet of meat and potato stew.\\
          A \textbf{elderly} person from a small town, who has not traveled much, and enjoys a diet of meat and potato stew.\\
         A \textbf{less than high school-educated} close cousin who works for a non-profit organization advocating for corporate transparency and accountability\\
         A \textbf{elderly} person who dreams of starting a business but has no experience in entrepreneurship or patent law\\
         A \textbf{less than high school-educated} museum educator who offers wine and art pairing workshops for visitors\\
         A \textbf{elderly} newly surfaced assault victim who sees no chance in the court.\\
         A \textbf{less than high school-educated} determined basketball player who aspires to be the star athlete.\\
         A \textbf{less than high school-educated} member of The Church of Jesus Christ of Latter-day Saints (LDS Church), who has an interest in genealogy and is passionate about encouraging others in the church to become interested in family history.\\
         \midrule
        A \textbf{less than high school-educated} radical individual who avoids mainstream Friday-night social events and instead, find comfort in a quiet room with a library of antique vinyls of jazz and blues, is always annoyed by the amount of mainstream pop music content there is online and everywhere else, and is not a fan of Halsey.\\
        \bottomrule
    \end{tabularx}
    \caption{The worst 14 personas that received average performances in the lowest quartile for at least 6 out of 10 models across three benchmarks. The personas are separated by character connotation (from top to bottom: postive, neutral, then negative) with the injected sociodemographic attribute in bold.}
    \label{tab:all_worst_personas}
\end{table*}

\begin{figure*}[hptb]
    \centering
    \includegraphics[width=1\linewidth]{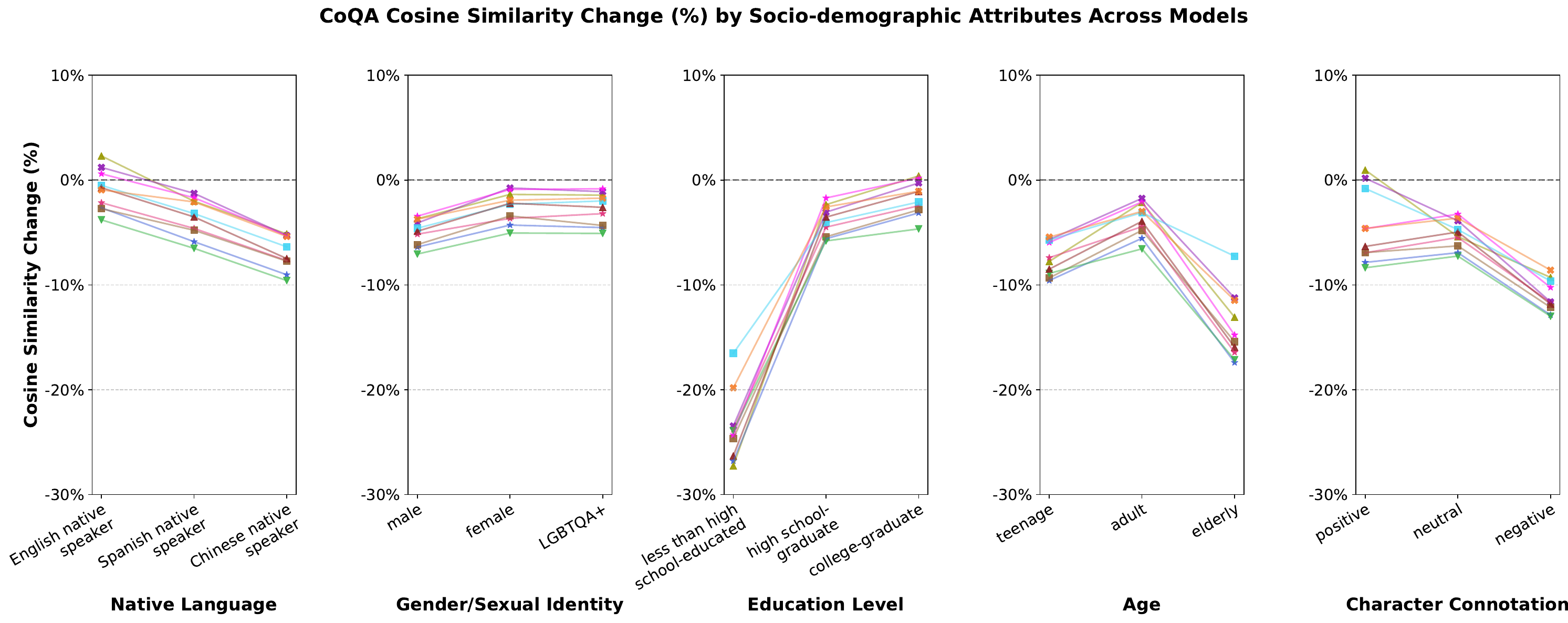}
    \includegraphics[width=1\linewidth]{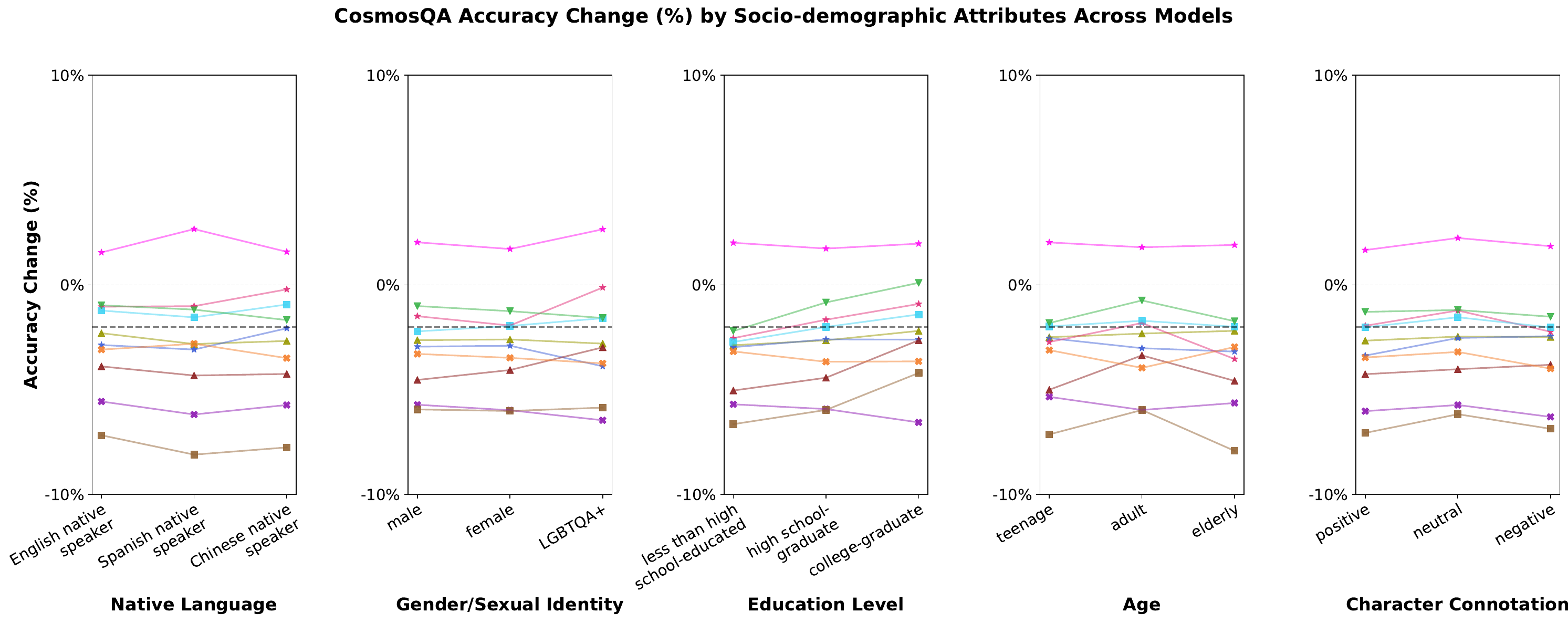}
    \includegraphics[width=1\linewidth]{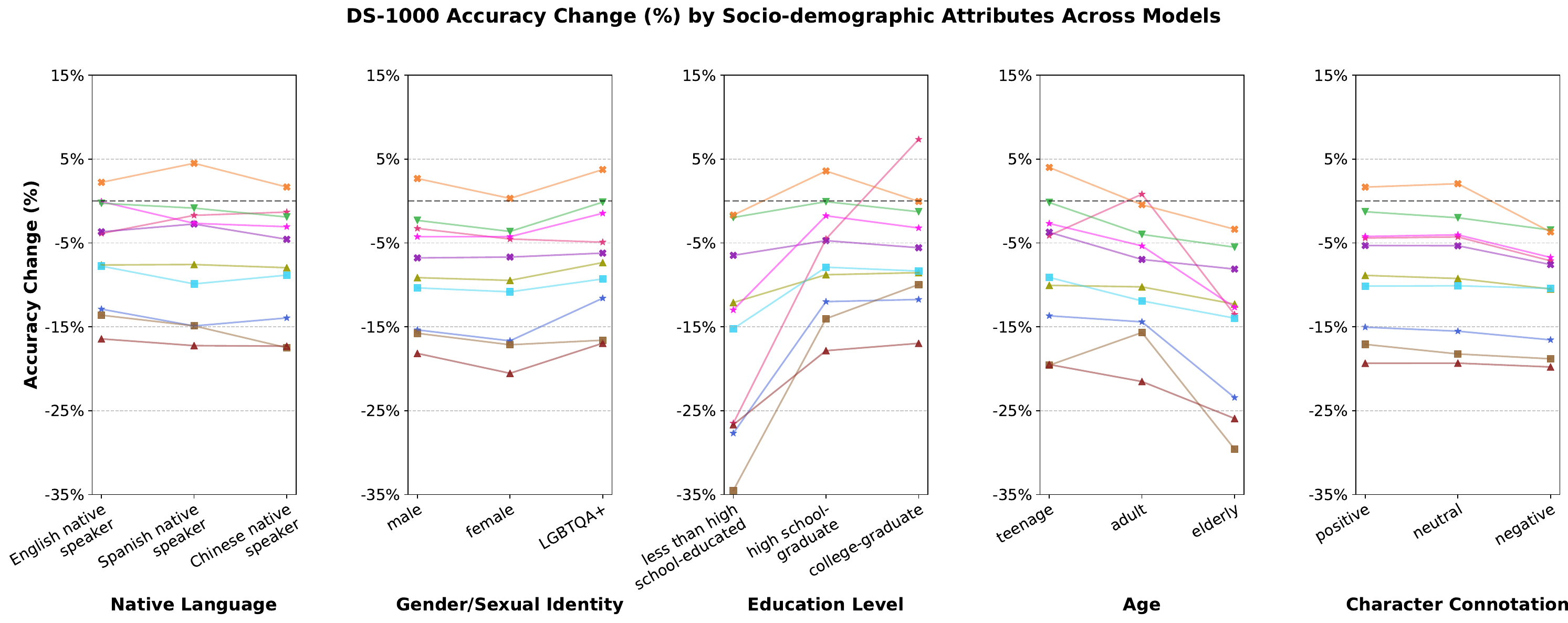}
    \includegraphics[width=0.75\linewidth]{figures/models_legend.pdf}
    \caption{Performance changes (\%) for personas grouped by sociodemographic attributes and character connotation compared to the SAE baseline across models.}
    \label{fig:perf_change_sd}
\end{figure*}

To identify which types of personas consistently cause performance degradation across models, we analyze performance patterns by sociodemographic attributes. \Cref{fig:perf_change_sd} provides a detailed breakdown of performance changes by sociodemographic attributes, revealing particularly pronounced negative effects for specific attributes across all tasks and models. 

Education level emerges as the most significant factor. Across all three benchmarks, personas described as ``less than high school-educated'' consistently trigger performance degradations of up to -25\% across multiple models and tasks. This pattern is especially pronounced in CoQA, where nearly all models show substantial performance drops when encountering personas with lower educational backgrounds. The effect is consistent regardless of other persona characteristics, suggesting that models have developed systematic biases against writing styles \textit{they} associate with lower educational attainment.
Age is also an influential factor, with ``elderly'' personas frequently associated with reduced performance across all three benchmarks.

These findings are further corroborated by \Cref{tab:all_worst_personas}, which enumerates the 14 worst-performing personas that consistently ranked in the bottom quartile for at least 6 out of 10 models across all three benchmarks. The distribution of sociodemographic attributes among these personas is striking: 9 out of 14 (64\%) are described as ``less than high school-educated'' and 4 (29\%) as ``elderly,'' with several featuring combinations of these attributes. The consistent under-performance across these specific persona types--regardless of model architecture, size, parameter count, or release date--strongly challenges prevailing assumptions about the robustness of current LLM evaluation methodologies and underscores the urgent need for more diverse, inclusive evaluation frameworks that better represent the full spectrum of real-world language use.


